\documentclass[10pt,twoside,journey]{IEEEtran}

\usepackage{amsfonts}
\usepackage{bm}
\usepackage{booktabs}
\usepackage{colortbl} 
\usepackage{xcolor}
\usepackage{subfigure}
\usepackage{multirow}
\usepackage{amssymb}
\usepackage{amsmath}
\usepackage{balance}
\usepackage{siunitx}

\ifCLASSOPTIONcompsoc
  \usepackage[nocompress]{cite}
\else
  \usepackage{cite}
\fi

\ifCLASSINFOpdf
   \usepackage[pdftex]{graphicx}
   \graphicspath{{../pdf/}{../jpeg/}}
   \DeclareGraphicsExtensions{.pdf,.jpeg,.png}
\else
   \usepackage[dvips]{graphicx}
   \graphicspath{{../eps/}}
   \DeclareGraphicsExtensions{.eps}
\fi

\usepackage{algorithm}  
\usepackage{algorithmic}
\usepackage{array}
\ifCLASSOPTIONcompsoc
 \usepackage[caption=false,font=footnotesize,labelfont=sf,textfont=sf]{subfig}
\else
 \usepackage[caption=false,font=footnotesize]{subfig}
\fi
\ifCLASSOPTIONcaptionsoff
 \usepackage[nomarkers]{endfloat}
\let\MYoriglatexcaption\caption
\renewcommand{\caption}[2][\relax]{\MYoriglatexcaption[#2]{#2}}
\fi
\let\MYorigsubfloat\subfloat
\renewcommand{\subfloat}[2][\relax]{\MYorigsubfloat[]{#2}}
\usepackage{url}

\begin{document}

\title{A Multi-Granularity Self-Guiding Graph Diffusion Model for Predicting Private Car Activity}

\author{
    Bo~Liu,
    Tong~Li,~\IEEEmembership{Member,~IEEE},
    Zhu~Xiao,~\IEEEmembership{Senior Member,~IEEE},~Wenjie~Chen,\\~Yongqiang Hei, Jingtao Ding,~Guo~Chen,~\IEEEmembership{Senior Member,~IEEE}
    
    \IEEEcompsocitemizethanks{\IEEEcompsocthanksitem Manuscript received XX XX 2025; revised XX XX 2025; accepted XX XX 2025.~This work was supported in part by the National Natural Science Foundation of China under Grant U24A20247, Grant 62471277 and Grant 62272152, in part by the Young Elite Scientists Sponsorship Program by CAST under Grant ZB2025-293,~in part by in part by the Hunan Natural Science Foundation of China under Grant 2025JJ50351, and in part by the Shenzhen Science and Technology Program under Grant JCYJ20250604185802004.~\textit{(Corresponding authors:~T.~Li, Z.~Xiao and W.~Chen.)}    
    \IEEEcompsocthanksitem B.~Liu, T.~Li, G. Chen, and Z.~Xiao are with the College of Computer Science and Electronic Engineering, Hunan University, Changsha 410082, China, and also with the Shenzhen Research Institute, Hunan University, Shenzhen 518055, China. Email: ~\{tliay,~zhxiao,~guochen\}@hnu.edu.cn.
    \IEEEcompsocthanksitem W.~Chen is with College of Economics and Management, Central South University of Forestry and Technology,  Changsha, Hunan 410004, China.~Email:~t20142193@csuft.edu.cn.
    \IEEEcompsocthanksitem Y.~Hei is with State Key Laboratory of Integrated Services Network (ISN), Xidian University, Xi'an 710071, China.~Email:~yqhei@mail.xidian.edu.cn.   
    \IEEEcompsocthanksitem J.~Ding is with Department of Electronic Engineering, Tsinghua University, Beijing, China.~Email:~dingjt15@tsinghua.org.cn.}      
}

\IEEEtitleabstractindextext{%
\begin{abstract}
Transportation networks have become a vital component of city infrastructure, with private cars accounting for the largest share of urban transportation. Private car activity reflects individualized travel behaviors, such as Arrive-Stay-Leave (ASL) sequences, which are critical for effective traffic management, parking optimization, and urban planning.
In this study, we propose MAEDE (Multi-grAnularity sElf-guiding Diffusion graph modEl) for probabilistic forecasting of regional private car activity. MAEDE employs a conditional diffusion model as its backbone, leveraging the strengths of diffusion models in probabilistic modeling and their ability to capture complex joint distributions. Additionally, MAEDE incorporates satellite imagery to capture environmental contexts using an LLM-enhanced contrastive learning module. In this module, the LLM generates textual descriptions related to private car activity for each image, and the contrastive learning component aligns these textual and visual representations, effectively encoding critical urban features.
To address both long-term trends and short-term fluctuations in private car activity, MAEDE uses a multi-granularity self-guiding mechanism. This mechanism utilizes coarse-grained data at various granularity levels as targets to guide the denoising process, enabling the simultaneous capture of high-level trends and fine-grained details.
Extensive experiments on real-world private car trip datasets demonstrate that MAEDE outperforms state-of-the-art baselines, achieving improvements of over $6.8\%$ in Root Mean Squared Error (RMSE) and $14.8\%$ in Continuous Ranked Probability Score (CRPS). Cross-city evaluations further validate MAEDE's robustness and generalization across diverse urban environments. \textcolor{black}{The code of MAEDE is available at: https://github.com/BoLiupro/MAEDE.}
\end{abstract}

}

\maketitle

\IEEEdisplaynontitleabstractindextext

\IEEEpeerreviewmaketitle

\section{Introduction}

With decades of global urbanization, transportation networks have become a critical and fundamental part of city infrastructure, facilitating efficient interactions among citizens and supporting the smooth exchange of goods and services~\cite{diao2021impacts, zhang2022cityneuro}. Private cars, which account for the largest share of urban transportation, play a significant role in this system. According to recent statistics~\cite{yiche2024, NBS2021}, over 90\% of households in the United States own at least one private car, while nearly 88.6\% of urban automobiles in China are private cars. Unlike public vehicles, such as buses and taxis, which serve broader communities, private cars are inherently tied to individual citizens and are deeply embedded in the daily routines of their owners~\cite{hu2023spatio, xiao2024exploring}, making them more reflective of human activities and personal needs. Analyzing private car activity patterns provides valuable insights that can drive data-informed decision-making, enabling more efficient traffic management, more innovative infrastructure planning, and promoting sustainable urban development. 

Private car activity refers to the movement of private vehicles, which can be characterized through Arrive-Stay-Leave (ASL) sequences~\cite{xiaoPredictingUrbanRegion2023, fang2023learning, li2019driving}. `Arrive' represents a vehicle's arrival at its destination, marking a regional transition and signaling the start of an interaction between the vehicle and the location. 'Stay' refers to the duration a vehicle remains parked at a location, with the length of stay influenced by factors such as the region's functionality and the driver's needs, e.g., whether the vehicle is parked for shopping, work, or leisure purposes. `Leave' marks the completion of the interaction as the vehicle departs and starts the next trip. Regional private car activity, as a whole, reflects the dynamic movement patterns of vehicles within a given area. Understanding these patterns is essential for applications such as predicting parking demand~\cite{xiao2023parking}, assessing the regional functions~\cite{zhang2021multi}, and optimizing traffic management strategies. 

Previous studies have extensively analyzed regional transportation mobility using datasets derived from public transportation systems such as metro, bus, and taxi services. For example, Yan et al.~\cite{yan2024bus} used IC card data to investigate bus-to-subway transfer patterns. Zhao et al.~\cite{zhao2022urban} developed planning strategies to integrate urban rail systems with conventional bus services, aiming to improve the efficiency of public transportation. Feng et al.~\cite{feng2017role} focused on the role of shared transportation in urban development, highlighting its influence on modern transit dynamics.
While these studies provide valuable insights, public transportation datasets have certain limitations, particularly in capturing stationary behavior and accurately reflecting individual trip intentions. Taxis, for instance, are predominantly demand-driven and continuously move between pick-up and drop-off points; however, they do not provide detailed information about vehicle stops. Similarly, buses operate on predetermined schedules and fixed routes, which makes them less indicative of individualized travel behaviors.
In contrast, analyzing private car activity presents a unique opportunity to capture stationary behavior, providing deeper insights into critical urban issues such as the attractiveness of commercial districts and parking shortages. Predicting private car activity at the regional level can enhance urban planning, inform better infrastructure design, and help reduce traffic congestion and emissions~\cite{speizer2024integrated}.

Accurately forecasting regional private car activity presents substantial challenges due to its inherent uncertainty, environmental complexity, and multi-scale temporal dynamics.
\textbf{Firstly}, private car activity is highly stochastic~\cite{li2014stochastic, liu2025vehicle}, shaped by static environmental factors and dynamic individual decision-making processes~\cite{jin2022survey, zhang2022toward}. As a result, regional private car activity exhibits considerable variability and inherent uncertainty. Existing studies that rely on deterministic models often fail to capture these fluctuations accurately, primarily due to the absence of proper probabilistic distribution-based modeling. Accurately modeling such variability and quantifying prediction uncertainty is essential for improving the reliability and robustness of predictions.
\textbf{Secondly}, static urban environments, such as road networks, building layouts, and regional functions, play a critical role in shaping private car activity~\cite{zhao2016urban, yuan2014discovering, xiao2025llm}. For instance, commercial districts typically experience higher arrival and stay rates during business hours, whereas residential areas see increased activity during mornings and evenings due to commuting and daily routines. Existing studies primarily use point-of-interest (POI) data~\cite{chen2018tripimputor, zhang2023deep} to represent environmental factors. However, POI distributions often fail to capture the intricate spatial interplay between urban form and car behavior, such as road connectivity, the geometry of building distributions, and mixed land uses. Accurately modeling these environmental contexts and extracting meaningful static spatial insights remain critical challenges. 
\textbf{Thirdly}, regional private car activity exhibits multi-granularity temporal patterns~\cite{liu2021understanding, wan2018smopat, cao2007discovery}, which span different time scales. Importantly, such temporal patterns are not uniform across space: some regions primarily follow daily rhythms, such as rush hour peaks driven by commuting, while others exhibit dominant weekly fluctuations, for example, increased visits to recreational areas on weekends~\cite{wang2019revealing}. Moreover, multiple temporal regularities often influence individual regions simultaneously, resulting in overlapping trends across different time scales. Traditional approaches are often limited in handling the different temporal scales of mobility patterns, making it difficult to capture both gradual trends and transient changes. Capturing multi-scale temporal dynamics under different periodicities remains a challenge in mobility forecasting.

To address these challenges, we propose a Multi-grAnularity sElf-guiding graph Diffusion modEl (MAEDE) for probabilistically forecasting regional private car activity. 
Specifically, MAEDE employs a conditional diffusion model as its predictive backbone, leveraging the strengths of diffusion models in probabilistic modeling and capturing complex joint distributions. By progressively denoising data and conditioning on both static environmental factors and dynamic historical temporal observations, MAEDE effectively captures regional activity fluctuations. Such a design enables it to generate robust, uncertainty-aware predictions, successfully addressing the \textbf{first} challenge. 

{
Furthermore, MAEDE utilizes satellite imagery to capture detailed environmental contexts. Compared to traditional POI data, satellite imagery offers abundant spatial and textual information, including road layouts, building geometries, and green spaces~\cite{li2023learning}. To extract static environmental features from satellite imagery, MAEDE integrates a Large-Language-Model (LLM)-enhanced Contrastive Language-Image Pre-training (CLIP) module. The LLM is prompted to generate private car activity-related textual descriptions for each satellite image. The CLIP module employs contrastive learning on these language-image pairs, aligning textual and visual representations to encode critical urban features, including road structures, land use, and urban functionality. We design a graph self-guiding encoder that fuses multi-granularity time-series input with multimodal environmental context to extract integrated representations, effectively capturing internal spatio-temporal dependencies within private car activities and the external influences of urban environments across multiple granularities. This process allows the model to develop a comprehensive contextual understanding of static environmental factors that influence private car behavior, effectively addressing the \textbf{second} challenge. 
Moreover, to effectively capture both long-term trends and short-term fluctuations in private car activity, MAEDE introduces a soft attention mechanism that computes attention weights for hidden representations at each granularity based on contextual relevance, adaptively emphasizing their complementary contributions to produce refined multi-granular representations. We design a multi-granularity diffusion module that employs a shared variance schedule to ensure that coarse-grained information dominates the early diffusion stages, while finer-grained details are progressively refined in later stages to output the final prediction through a coarse-to-fine self-guiding denoising process. The multi-granularity design allows the model to learn long-term evolution trends and short-term fluctuations simultaneously, addressing the \textbf{third} challenge.
}

In this framework, temporal activity data is divided into multiple granularities. Encoders are trained to separately predict the following points at each granularity, extracting hidden states for different levels of granularity. These hidden states are then adaptively fused using an attention module and subsequently used as a condition to guide the learning process of the diffusion models. During the denoising process, the coarse-grained data constrain latent states, preserving long-term trends. This strategy encourages the generation of coarser features in early stages of diffusion, while facilitating the recovery of finer-grained features in later stages.

\begin{figure*}[tb]
    \centering
    \vspace{-2mm}
    \includegraphics[width=0.7\linewidth]{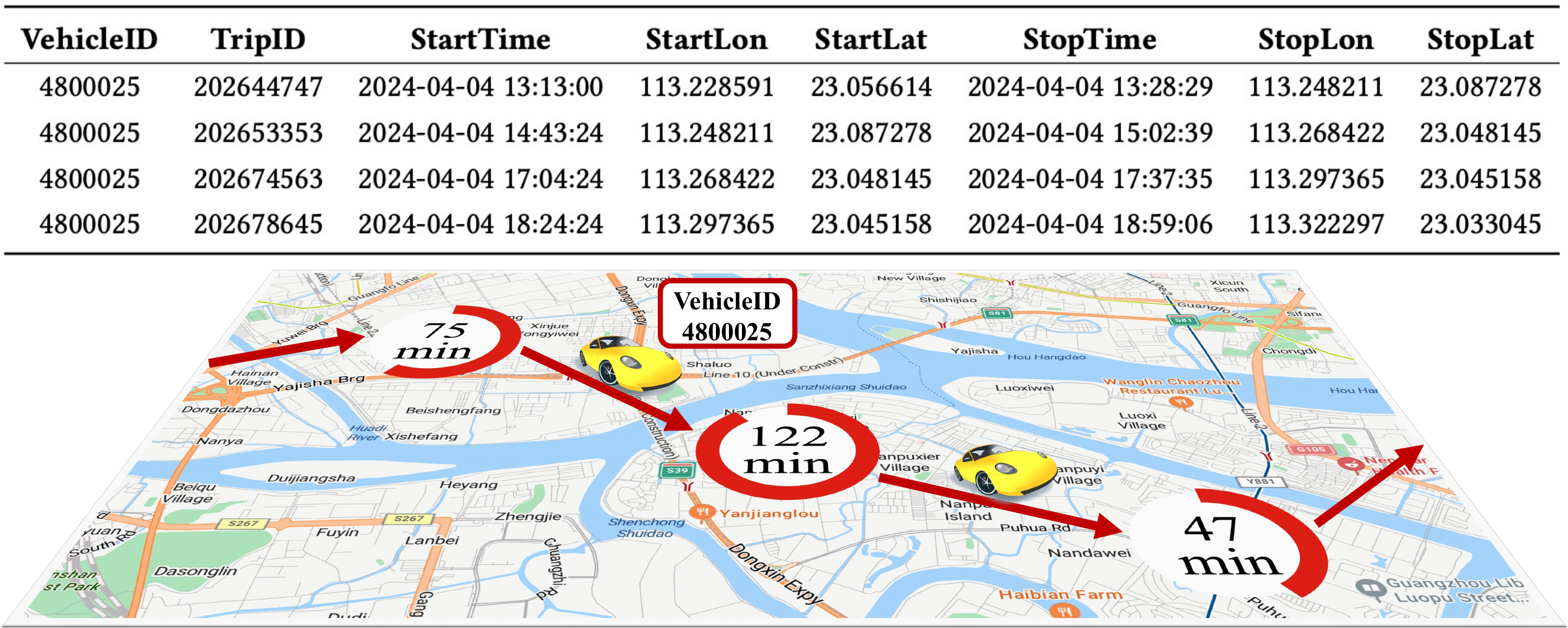}
    \vspace{-2mm}
    \caption{Private car trip dataset illustration. Each record consists of VehicleID, TripID, StartTime, StartLon, StartLat, StopTime, StopLon, and StopLat. Duration of driving and parking can be calculated by StartTime and StopTime.}
    \vspace{-2mm}
    \label{fig:preliminaries_dataset}
\end{figure*}

The main contributions can be summarized as follows:

$\bullet$ We investigate the problem of predicting regional private car activity based on historical observations while incorporating external environmental factors. Additionally, we highlight the unique advantages of studying private car activity, which provides crucial stationary information and captures the interplay between trip behavior and regional dynamics.

$\bullet$ We propose MAEDE. MAEDE employs a conditional diffusion model as the backbone for probabilistic modeling, utilizes an LLM-enhanced CLIP to extract static environmental features, and incorporates a multi-granularity self-guiding mechanism to capture both coarse-grained trends and finer-grained features jointly. MAEDE offers a robust, uncertainty-aware solution for predicting regional private car activity.

{$\bullet$ We conduct extensive experiments on private car trip datasets collected from multiple cities, demonstrating that MAEDE outperforms state-of-the-art baselines by around $6.8\%$ in RMSE and $14.8\%$ in CRPS. Ablation studies validate the effectiveness of each component, while cross-city evaluations showcase the robustness and generalization of MAEDE across diverse urban environments.}
\section{Preliminaries}
\subsection{Private Car Activity}
\label{private_car_activity}

\begin{figure*}[tb]
    \centering
    \vspace{-2mm}
    \includegraphics[width=1\linewidth]{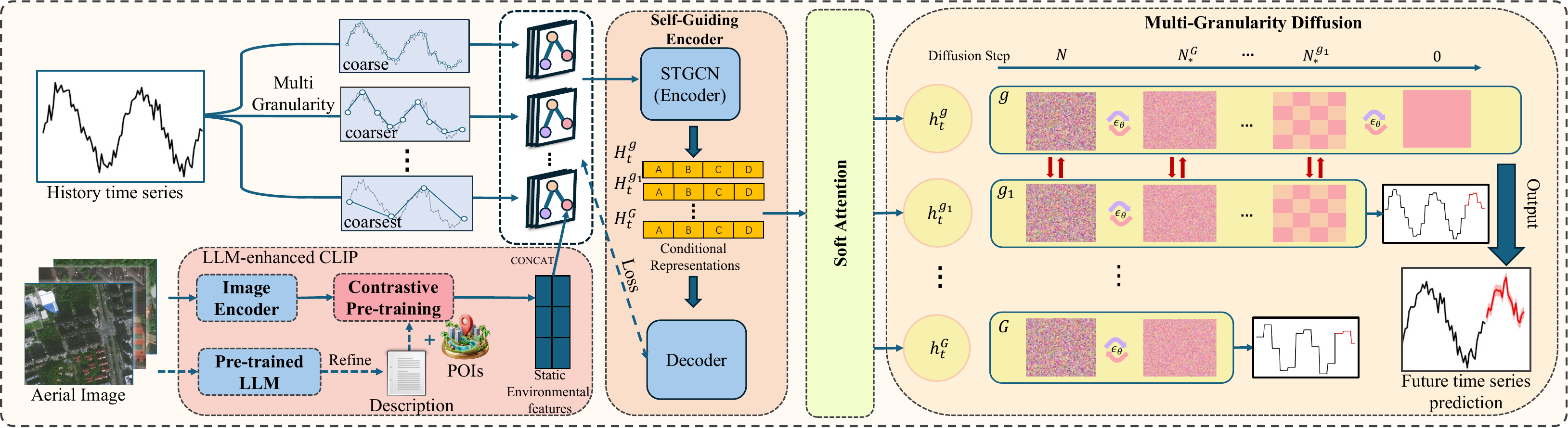}
    \vspace{-2mm}
    \caption{Framework of Multi-grAnularity sElf-guiding graph Diffusion Model (MAEDE). MAEDE consists of four major components: (1) LLM-enhanced CLIP for capturing environmental conditions, (2) self-guiding encoder for automatically extracting conditional representations, (3) features soft attention fusing mechanism for fusing multi-granularity features, and (4)multi-granularity diffusion model for generating probabilistic forecasts.}
    \vspace{-2mm}
    \label{fig:framework}
\end{figure*}

Private car activity provides a fine-grained characterization of regional traffic, capturing not only the geographical locations of private cars but also their stationary durations. Specifically, regional private car activity can be described through the ASL behavior, a continuous sequence including arrival at the fine-grained destination, a stationary period, and subsequent departure to the next destination. 

Let $Tr^v =\{d^v_1, d^v_2, \dots, d^v_t\}$ represent the trip point sequence of private car $v$ observed over a sequence of time intervals $t \in [0, \dots, T]$, where each point is denoted as  $d^v_t=\{lon^v_t,lat^v_t,\delta^v_t\}$ represent each trip point. $lon^v_t$ and $lat^v_t$ denote the location, while $\delta^v_t$  denotes the stay duration. Note that~$\delta^v_t$ can be computed based on the vehicle's arrival and leave times. However, $\delta^v_t$ exceeding a set observation time window $\Delta T$ will be truncated, and excessively long $\delta^v_t$ is excluded, as the vehicle is considered inactive. In this case, given a partitioned urban region $r\in \{r_1,r_2,\cdots,r_R\}$, where $R$ denotes the total number of regions in the city, the ASL behavior within a region can be decomposed into: $S_{r,t}^{in}$, $S_{r,t}^{out}$,  $S_{r,t}^{remain}$, denoting arriving to a region from another, leaving the previous region to another, and remaining in the same region, respectively. These compositions from ASL behaviors are formulated:

\begin{equation}
    S_{r,t}^{in} = \{~d^v_t~|~d^v_{t-1} \notin r \wedge d^v_t \in r\},
\end{equation}
\begin{equation}
    S_{r,t}^{out} = \{~d^v_t~|~ d^v_{t-1} \in r\wedge d^v_t \notin r\},
\end{equation}
\begin{equation}
    S_{r,t}^{remain} ~= \{~d^v_t~| ~d^v_{t-1} \in r \wedge d^v_t \in r\},
\end{equation}
where $d_t^v \in r$ represents the trip points of vehicle $v$ located in the region $r$ in observed time window. Notably, instances of overnight parking or extended inactivity beyond the typical observation window are excluded. In this context, regional private car activity can be numerically represented by formulating ASL behaviors:
\begin{equation}
    X_{r,t} = \sum \delta^v_t+\rho|S^{in}_{r,t}+S^{remain}_{r,t}|,
\end{equation}
where $X_{r,t}$ denotes the private car activity of region $r$ in the time window $t$, $\delta^v_t$ denotes the stay period of  trip point $\quad d_t^v \in S_{r,t}^{in} \cup S_{r,t}^{remain}$. $\rho$ is the weight to integrate stationary information with dynamic mobility features. $S^{out}_{r,t}$ is not included in calculation of  $X_{r,t}$ since it is considered as $S^{out}_{r,t'}$ of other regions. Integrated with regional traffic data and stationary information, regional private car activity is particularly valuable for optimizing transportation planning, parking management, and policy development, offering a more comprehensive understanding of vehicle behavior.

Private car trip dataset includes trip data from major cities in 34 provincial-level administrative regions in China, spanning from January 2022 to August 2024. The trip data is collected by in-vehicle GPS, achieving road-level positioning accuracy, and has undergone desensitization processing to ensure privacy protection. Fig.~\ref{fig:preliminaries_dataset} illustrates samples of trip records. VehicleID and TripID identify unique private car and trip records, while other variables describe how a trip is organized. 

\subsection{Conditional Diffusion Model}
Denoising Diffusion Probabilistic Models (DDPMs)~\cite{ho2020denoising, nichol2021improved} are generative models that learn to reverse a predefined noising process, progressively transforming simple noise distributions into complex data distributions through denoising steps. The conditional Denoising Diffusion Probabilistic Model (Conditional DDPM)~\cite{song2021scorebased} extends the standard DDPM framework by incorporating additional conditioning information $c$ to guide the generative process.

Predicting regional private car activity can be modeled using a conditional diffusion model to generate a probabilistic forecasting task. The condition $c$ encompasses multi-granularity features derived from various data sources. The forward process remains a fixed Markovian Gaussian process, where noise is gradually added to the original data $x_0$, but now conditioned on $c$:

\begin{equation}
    q(x_{1:K} | x_0, c) = \prod_{k=1}^{K} q(x_k | x_{k-1}, c),
\end{equation}
\begin{equation}
    q(x_k | x_{k-1}, c) = \mathcal{N}(x_k; \sqrt{1 - \beta_k} x_{k-1} + f_c(c, k), \beta_k \zeta),
\end{equation}
where \( x_0 \) represents the original data sample, and \( x_k \) denotes the noisy data at step \( k \), with \( K \) being the total number of diffusion steps. \( \beta_k \) controls the noise variance $\zeta$ at step \( k \), and \( f_c(c, k) \) represents an additional conditioning function that incorporates the external input \( c \) into the diffusion process. The reverse process involves learning to progressively remove noise while incorporating \( c \) as an additional conditioning factor:
\begin{equation}
    p_\theta(x_{0:K} | c) = p(x_K) \prod_{k=1}^{K}\mathcal{N}(x_{k-1}; \mu_\theta(x_k, k, c), \sigma_\theta(x_k, k, c)\zeta),
\end{equation}
where \( \mu_\theta(x_k, k, c) \) and \( \sigma_\theta(x_k, k, c) \) are parameterized by a neural network that takes both the noisy input \( x_k \) and the conditioning variable \( c \), allowing for controlled denoising. 

This objective encourages the network to correctly predict the noise added at each time step while leveraging conditioning information \( c \). The training loss is:
\begin{equation}
    \mathcal{L}(\theta) = \mathbb{E}_{x_0, \epsilon, k, c} \Big\| \epsilon - \epsilon_\theta (x_k, k, c) \Big\|_2^2.
\end{equation}

\subsection{Problem Definition}
Using historical observations of regional private car activity, urban spatial data, and environmental factors such as satellite imagery and Points of Interest (POI) information, we aim to predict future private car activity within a specified time window. Mathematically, let $X_{r,0:t_0-1} = \{x_{r,1}, x_{r,2}, \dots, x_{r,t_0-1}\}$ represent the private car activity observed in the region $r$ over time steps $ t \in [1, T]$. Let $A$ denote the adjacency information of regions (e.g., spatial or connectivity relationships), and let $C$ represent the environmental conditions (e.g., satellite data, POI distributions).
Our goal is to build a predictive model $F$ that takes the input data $X_{r,t}$,~$A$, and $C$ to generate accurate forecasts of future private car activity, $X_{r,t_0:T}$ for region $r$. This predictive task can be formally expressed as:
\begin{equation}
    \hat{X}_{r,t_0:T}=F\left(X_{0:t_0-1}, A, C\right),
\end{equation}
where $\hat{X}_{r,t_0:T}$ represents the predicted private car activity for the future time steps $t \in [t_0, T]$.

\section{MAEDE}

\subsection{Overview}

A Multi-grAnularity sElf-guiding graph Diffusion Model (MAEDE) is proposed to enhance probabilistic forecasting in urban mobility analysis. MAEDE consists of four major components: (1) a pre-trained LLM-enhanced CLIP encoder; (2) a self-guiding feature encoder;  (3) a feature soft attention fusion mechanism; and (4) a multi-granularity diffusion model. The framework of our proposed model is presented in Fig.~\ref{fig:framework}. The process begins with an LLM-enhanced pre-trained CLIP model, which captures static environmental features from satellite imagery. A self-guiding encoder then processes these extracted features and the multi-granularity spatiotemporal graph to generate conditional representations. The multi-granularity conditional representations are subsequently fused through a soft attention module, which dynamically emphasizes the most relevant information. Finally, the fused features are fed into a multi-granularity conditional diffusion model, generating probabilistic forecasts.

\subsection{LLM-enhanced CLIP}
Understanding the static environment is crucial for modeling private car activity, as urban environmental factors, such as road networks, land-use patterns, and infrastructure, significantly influence mobility behavior~\cite{xi2022beyond, long2023practical, li2022predicting}. However, existing methods typically rely on POI distributions or raw imagery, which either lack spatial structural cues or semantic interpretability. This motivates the need to integrate POI semantics with visual features for more expressive representations of the environment.

We propose an LLM-enhanced CLIP module based on a contrastive learning framework. The module aligns satellite imagery with semantically enriched POI descriptions to learn environmental static representations. Specifically, an image encoder processes high-resolution satellite images to extract spatial textures, while a text encoder encodes natural language descriptions derived from POI data. These descriptions are generated by a pre-trained LLaMA-Adapter V2, which transforms structured POI tags into fluent and informative sentences that contextualize regional functional semantics. Text and image encoders are jointly trained using a contrastive objective to align the two modalities. The loss is defined as:
\begin{equation}
\mathcal{L}_{\text{CLIP}} = \frac{1}{2} \left( \mathcal{L}_{\text{img} \rightarrow \text{text}} + \mathcal{L}_{\text{text} \rightarrow \text{img}} \right),
\end{equation}
\[
\mathcal{L}_{\text{img} \rightarrow \text{text}} = -\sum_{i=1}^{R'} \log \frac{\exp\left(\text{sim}(E_I(I_i), E_T(T_i)) / \tau\right)}{\sum_{j=1}^{R} \exp\left(\text{sim}(E_I(I_i), E_T(T_j)) / \tau\right)},
\]
\[
\mathcal{L}_{\text{text} \rightarrow \text{img}} = -\sum_{i=1}^{R'} \log \frac{\exp\left(\text{sim}(E_T(T_i), E_I(I_i)) / \tau\right)}{\sum_{j=1}^{R} \exp\left(\text{sim}(E_T(T_i), E_I(I_j)) / \tau\right)},
\]
where $E_I(\cdot)$ and $E_T(\cdot)$ denote the image and text encoders, $\text{sim}(\cdot,\cdot)$ represents cosine similarity, and $\tau$ is a temperature parameter. $R'$ denotes the number of regions. This contrastive loss brings matched image-text pairs closer in the embedding space while pushing apart mismatched pairs, enabling effective cross-modal fusion and facilitating the extraction of rich, semantically grounded environmental representations.

\subsection{Self-Guiding Encoder}
\label{sefl_guiding_encoder}
{
Conditional information contains crucial spatio-temporal dependencies that are beneficial for various learning tasks. Since conditional information remains consistent across different inference samples and generalizes across tasks, it should be pre-extracted effectively to enhance the performance and stability of subsequent denoising models~\cite{wenDiffSTGProbabilisticSpatioTemporal2024, liu2023pristi, hu2024towards}.}

\begin{figure}[tb]
    \centering
    \vspace{-2mm}
    \includegraphics[width=1\linewidth]{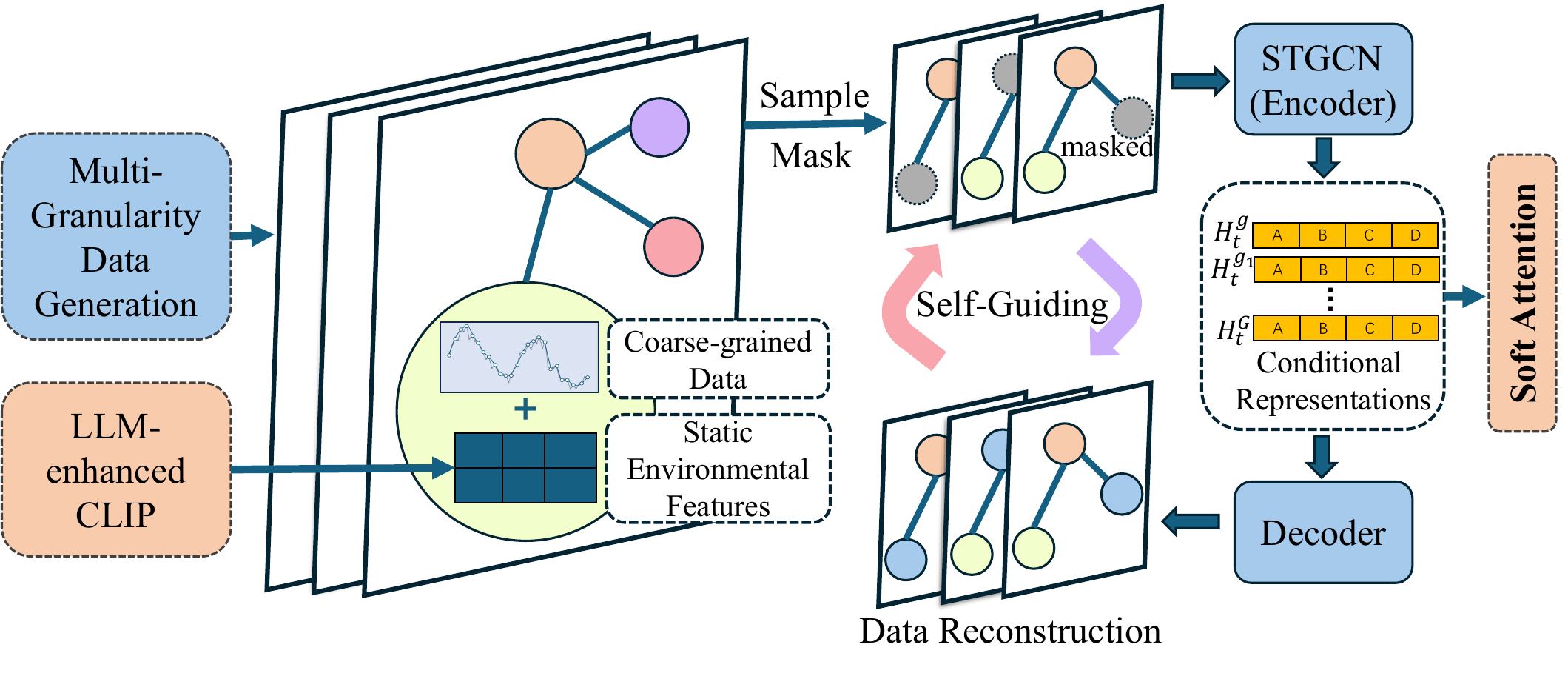}
    \vspace{-2mm}
    \caption{Details of Self-Guiding Encoder. STGCN Encoder extracts features from processed spatiotemporal topology, where each node contains coarse-grained data fused with environmental conditions. Next, attention-based denoising decoders generate predictions by leveraging learned patterns through self-guiding.}
    \vspace{-2mm}
    \label{fig:self guiding encoder}
\end{figure}

We propose a pre-trained self-guiding encoder that derives latent conditional representations from multi-granularity inputs via an unsupervised autoencoding framework, as illustrated in Fig.~\ref{fig:self guiding encoder}. 
\textcolor{black}{
The input is organized as a spatiotemporal topology graph, where each node is associated with coarse-grained temporal signals and static environmental features extracted by the LLM-enhanced CLIP module. During training, we follow Masked Autoencoder and randomly mask node signals, such that only a subset of node features remains visible to the model.
The self-guiding encoder is then trained to reconstruct the masked node signals based on the partially observed graph and its relational structure, forming an implicit reconstruction objective in the spirit of masked autoencoding. This training scheme encourages the encoder to infer missing information by exploiting structural dependencies and contextual cues, rather than relying on complete or city-specific observations.
As a result, the learned node representations capture stable functional semantics and interaction patterns that are shared across different urban environments. These representations can therefore be reused to guide downstream diffusion modeling in unseen cities, providing an effective inductive bias for cross-city transfer without requiring retraining of the encoder.}

{
The encoder consists of a stack of four spatiotemporal graph neural network (STGNN) layers based on the Gated Graph WaveNet architecture. Each layer comprises a dilated temporal convolution to capture long-range temporal dependencies, followed by a graph convolution to model spatial correlations. Residual and skip connections are incorporated throughout to ensure stable feature propagation.}

\subsection{Soft Attention}
{
One of the key challenges in our framework lies in effectively fusing the multi-granular conditional representations extracted by the Self-Guiding Encoder. These representations, derived from different temporal granularities, encode complementary information, and their coherent integration is crucial for preserving hierarchical spatio-temporal dependencies.}

To address this, we employ a soft attention-based fusion mechanism~\cite{li2022predicting}, which adaptively assigns importance weights to different granularities based on their contextual relevance. After the encoder produces a set of conditional representations $H_r = \{H_r^{g_1}, H_r^{g_2}, \cdots, H_r^{g_G}\}$ for region $r$ at $G$ granularities, we compute attention weights $\alpha_{g_i}$ for each granularity using a learnable gating function:

\begin{equation}
    h^{g_i}_r = \alpha_{g_i} \cdot H_r,
\end{equation}
\[
    \alpha_{g_i} = {H_r}^T \cdot \sigma(W_1 \cdot H^{g_1}_r +\cdots + W_G\cdot H^G_r + b),\quad i = 1, 2, ..., G,
\]
where $\{W_1, W_2, \cdots, W_G\}$ are learnable projection weights for each granularity, $b$ is a shared bias term, and $\sigma(\cdot)$ denotes a non-linear activation function. The resulting fused embedding $h^{g_i}_r$ captures the weighted contribution of all granularities for a specific $g_i$. {Instead of directly injecting external conditions at each diffusion step, we jointly encode temporal patterns and environmental context, and the resulting fused vector serves as the initial hidden representation in the RNN backbone of the diffusion process. It encapsulates both temporal dynamics and multimodal environmental context, providing implicit conditioning throughout the denoising process.
}
{
This soft attention module is implemented as a two-layer feedforward network with layer normalization, dropout, and ReLU activations. The attention weights are further refined through a sigmoid gating mechanism, enabling the model to emphasize informative granularities while suppressing noisy signals selectively.}

\subsection{Multi-Granularity Conditional Diffusion}

To enhance performance in spatiotemporal forecasting, we introduce a Conditional Diffusion Model that leverages the inherent granularity levels within the data, utilizing given targets at intermediate diffusion steps to guide learning process. \textcolor{black}{Unlike most existing diffusion-based forecasting methods, which either operate at a single temporal granularity and treat diffusion as a purely noise-driven generative process conditioned on past observations, MAEDE introduces a multi-granularity self-guiding diffusion mechanism. The key intuition is to guide diffusion using multi-granularity temporal representations rather than single resolution signals. Coarse-grained diffusion processes capture global temporal patterns and long-term trends, while fine-grained diffusion focuses on localized and short-term dynamics. These diffusion processes are hierarchically coupled, such that coarse-grained diffusion provides guidance for fine-grained diffusion during generation. Through this hierarchical interaction across temporal granularities, the diffusion process is guided from coarse-scale temporal structures to fine-scale dynamics, enabling coherent and stable generation across multiple forecasting horizons rather than being driven by a single temporal resolution.}

\subsubsection{Multi-Granularity Data Generation}

Multiple-granularity data is constructed using a smoothing function over the original fine-grained series.{ Formally, supposing $f(\cdot)$ a simple moving-average function applied on non-overlapping windows and $s^g$ is the sliding window size associated with granularity $g$, the coarse-grained data at granularity $g$ is then obtained as $X^{(g)} = f(X^{(1)}, s^g)$.} The sliding windows are non-overlapping, and the obtained coarse-grained data is replicated $s^g$ times to align over the timeline $[1, T ]$. Each granularity level captures distinct temporal dependencies and trends, enabling a more comprehensive understanding of complex mobility patterns. 

\subsubsection{Multi-Granularity Guiding Diffusion}
{
To incorporate the coarse-grained information, we introduce an intermediate diffusion step $N^g_*$ at which the generated sample is expected to align with the coarse-grained data. Since direct likelihood estimation is intractable, we employ a variational lower bound:
\begin{equation}
    \log p_\theta(x^g) = \log \int p_{\theta}(x^g_{N^g_*}, x^g_{N^g_*+1}, ..., x^g_N) dx^g_{(N^g_*+1):N},
\end{equation}
where \(x^g\) denotes coarse-grained data, \(p_{\theta}(\cdot) \) denotes the probability distribution, \( \theta \) is the learnable parameters, \( N \) denotes the number of diffusion steps, and \( N^g_* \) specifies the intermediate step. To ensure the coarse-grained information influences the early stages of diffusion while finer-grained details are refined in later stages, we define a shared variance schedule:
\begin{equation}
    \alpha^g_n(N^g_*) =
    \begin{cases} 
        1, & n = 1, ..., N^g_* \\
        \alpha^1_n, & n = N^g_*+1, ..., N,
    \end{cases} 
\end{equation}
and $\{\beta_n^g\}_{n=1}^N = \{1 - \alpha_n^g\}_{n=1}^N$. 
Accordingly, define 
$a_n^g(N_*^g) = \prod_{k=1}^{n} \alpha_k^g$, 
and $b_n^g(N_*^g) = 1 - a_n^g(N_*^g)$. $N_\ast^g$ controls the temporal mapping of each granularity to the diffusion progress, 
while $\alpha_n^g(N_\ast^g)$ adjusts its relative weighting within the global denoising hierarchy. 
Coarse granularities dominate early diffusion stages (when $n \leq N_\ast^g$), 
while finer branches receive increasing noise injection $(1 - \alpha_n^g)$ as $n$ grows, 
enabling a stage-wise dominance from global trends to local variations.
}

\subsection{Self-Guiding Training}
The guidance loss function \( L^{(g)}(\theta) \) for the \( g \)-th granularity \( x^g_{n,t} \) at timestep \( t \) and diffusion step \( n \) can be expressed as:

\begin{equation}
    L^{(g)}{(\theta)} = \mathbb{E}_{\epsilon, x^g_{0,t}, n} \|\epsilon - \epsilon_{\theta}(\sqrt{a^g_n} x^g_{0,t} + \sqrt{b^g_n} \epsilon, n, h^g_{t-1})\|^2_2,
\end{equation}
where $h^g_{t-1}$ is the fused features from the Soft Attention.

To extend beyond a single level of guidance, we incorporate multi-granularity constraints in the diffusion process. To enforce the multi-granularity constraints across all levels, the final training objective is formulated as:
\begin{equation}
    L_{\text{final}} = \sum_{g=1}^{G} \omega_g \mathbb{E}_{\epsilon, x^g_{0,t}, n} \|\epsilon - \epsilon_{\theta}(x^g_{n,t}, n, h^g_{t-1})\|_2^2,
\end{equation}
where $\omega_g$ controls the contribution of each granularity. 

\section{Experiments}

{
In this section, we conduct experiments on the real-world private car dataset and compare the performance of our proposed method with various baselines. We also run extensive ablation studies to verify the effectiveness of individual modules in our method. Then, the generalization ability of our method is examined across multiple cities.  
}
\begin{table*}[t]
    \centering
    \footnotesize
    \caption{Statistics and Attributes of the Private Car Trip Dataset.}
    \label{tab:dataset_attr}
    \begin{tabular}{ccccccc}
        \toprule
        \textbf{City} & \textbf{\#Records} & \textbf{\#Vehicles} &
        \textbf{Time Span} & \textbf{Longitude Range} & \textbf{Latitude Range} \\
        \midrule
        Changsha  & 572{,}453 & 26{,}578 & 2022/12/01--2022/12/15 & 112.915--113.035 °E & 28.115--28.235 °N \\
        Guangzhou & 643{,}610 & 29{,}180 & 2024/01/01--2024/01/15 & 113.177--113.297 °E & 23.071--23.191 °N \\
        Shenzhen  & 655{,}896 & 27{,}245 & 2022/08/01--2022/08/15 & 113.927--114.047 °E & 22.566--22.686 °N \\
        \midrule
        \multicolumn{1}{c}{\textbf{Attribute}} &
        \multicolumn{2}{c}{\textbf{Description}} &
        \multicolumn{1}{c}{\textbf{Attribute}} &
        \multicolumn{2}{c}{\textbf{Description}} \\
        \midrule
        VehicleID & \multicolumn{2}{c}{Unique identifier for each private car.}
                  & TripID    & \multicolumn{2}{c}{Unique identifier for each trip record.} \\
        StartTime & \multicolumn{2}{c}{Timestamp indicating the start of the trip.}
                  & StopTime  & \multicolumn{2}{c}{Timestamp indicating the end of the trip.} \\
        StartLon  & \multicolumn{2}{c}{Longitude of the trip’s starting point.}
                  & StopLon   & \multicolumn{2}{c}{Longitude of the trip’s ending point.} \\
        StartLat  & \multicolumn{2}{c}{Latitude of the trip’s starting point.}
                  & StopLat   & \multicolumn{2}{c}{Latitude of the trip’s ending point.} \\
        \bottomrule
    \end{tabular}
\end{table*}

\subsection{Private Car Dataset}

{
Our study utilizes a comprehensive private car trip dataset~\cite{liu2025vehicle, xiao2021trajectory, xiao2024exploring} comprising 1,871,959 hourly records collected between March 1 and August 27, 2024. The data were obtained from three major Chinese cities: Changsha, Guangzhou, and Shenzhen. Table~\ref{tab:dataset_attr} summarizes the available attributes in the dataset, including trip start and stop times, spatial coordinates, and vehicle identifiers.
}

{
To model urban regional spatial structure, we constructed a topological graph by clustering grids based on trip patterns. Specifically, each city was divided into $25 \times 25$ grids. Trip points outside predefined urban boundaries were excluded. We focused on private car trips during morning (8:00–9:00) and evening (17:00–18:00) peak hours. To capture functional regions, DBSCAN~\cite{ester1996density} clustering was applied to trip-point coordinates with parameters $eps = 100 m$, $min\_samples = 25$; grid cells in the same cluster were merged into regions. For consistency, 45 regions per city were selected to construct the urban mobility graph.
Regional private car activity can then be extracted from raw trip data. Based on the StartTime and StopTime, we calculate the driving and parking durations for each trip, using a 60-minute observation window. In addition, records with a stationary duration of more than 4 hours were regarded as inactive (e.g., overnight parking) and removed, following empirical evidence that the average duration of parking in Chinese CBD areas is approximately 3.8 hours~\cite {mesfin2024urban}, which implies that more extended stays correspond to events of non-continuous mobility. By analyzing the spatial coordinates of trip origins and destinations, we infer movement trajectories and extract patterns of vehicle behavior. Regional private car activity was then quantified based on established formulations of aggregate trip frequency and duration within each region.
}
This study strictly adheres to data privacy and ethical standards. All data were collected with the informed consent of the users and full authorization from the data provider. Only anonymized, region-level aggregate data were shared with researchers, and no access to individual-level trajectories was granted.

\subsection{Satellite Imagery}

\begin{figure}[tb]
    \centering
    \vspace{-2mm}
    \subfigure[Office region]{\includegraphics[width=0.33\linewidth]{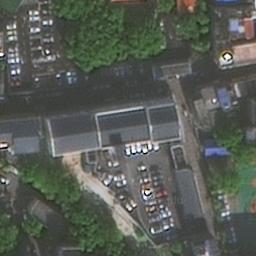}}
    \subfigure[School region]{\includegraphics[width=0.33\linewidth]{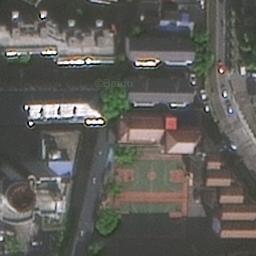}}\\
    \subfigure[Residential region]{\includegraphics[width=0.33\linewidth]{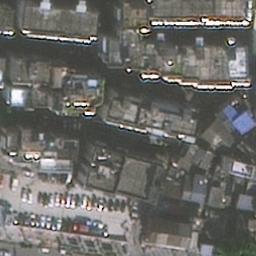}}
    \subfigure[Entertainment region]{\includegraphics[width=0.33\linewidth]{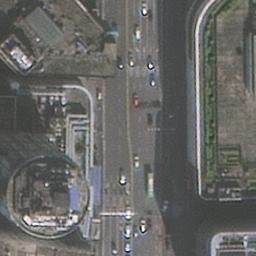}}
    \vspace{-2mm}
    \caption{Satellite images of different types of urban regions.}
    \vspace{-2mm}
    \label{fig:satellite_imagery}
\end{figure}

To effectively model static environmental context, we propose an LLM-enhanced CLIP module that employs an image encoder to extract visual features from high-resolution satellite imagery and aligns them with POI-based textual semantics. Specifically, we collect satellite imagery with the highest available zoom level (zoom = 19) from Baidu Map. The samples of satellite imagery in various regions are shown in Fig.~\ref{fig:satellite_imagery}. In total, we obtain 3,008 images from Changsha, 3,428 from Guangzhou, and 3,560 from Shenzhen, resulting in a diverse and representative dataset for learning environmental features. To construct text-image pairs, we employ a pre-trained LLM to convert structured POI  into natural language captions of each image. These pairs are used to train the contrastive alignment between modalities in the CLIP framework, facilitating fusion of spatial textures and functional semantics.

\subsection{Baselines}
\label{app:baselines}

We compare our method with the five SOTA baselines.

$\bullet$ \textbf{LSTM} (Long Short-Term Memory)~\cite{hochreiter1997lstm}. LSTM networks are a type of recurrent neural network designed to capture long-range temporal dependencies. They have been widely used for sequence modeling and, in this context, serve as both the generator and discriminator in a GAN framework for generating cellular traffic.

$\bullet$ \textbf{STGCN} (Spatio-Temporal Graph Convolutional Network)~\cite{yu2018stgcn}. STGCN integrates graph convolutions for modeling spatial relationships with temporal convolutions to capture dynamic temporal dependencies. It is applied in spatio-temporal forecasting by simultaneously learning spatial interactions and temporal evolution patterns.

$\bullet$ \textbf{TimeGrad}~\cite{rasulAutoregressiveDenoisingDiffusion}. By iteratively refining predictions through noise reduction, TimeGrad effectively captures temporal dependencies, making it suitable for probabilistic time series forecasting.

$\bullet$ \textbf{USTD} (Unified Spatio-Temporal Diffusion Model)~\cite{huUnifyingDiffusionModels2024}. USTD unifies spatio-temporal graph learning tasks under a probabilistic diffusion framework. It employs a shared spatio-temporal encoder with attention-based denoising networks, effectively capturing conditional dependencies for forecasting and spatial interpolation tasks.

$\bullet$ \textbf{DiffSTG} (Diffusion-based Spatio-Temporal Graph Network)~\cite{wenDiffSTGProbabilisticSpatioTemporal2024}. DiffSTG extends denoising diffusion probabilistic models to spatio-temporal graphs, offering a non-autoregressive framework for uncertainty-aware forecasting. Combining spatio-temporal learning with diffusion-based probabilistic modeling enhances robustness and stability in dynamic traffic prediction.

$\bullet$ \textbf{TACTiS} (Temporal-Attentive Context for Time Series)~\cite{tashiro2023tactis}. TACTiS models conditional distributions of multivariate time series by learning temporal context representations through transformer-based attention mechanisms. It enables fast, parallelizable sampling and uncertainty-aware forecasting by integrating diffusion modeling with temporal attention.

$\bullet$ \textbf{CSDI} (Conditional Score-based Diffusion for Imputation)~\cite{tashiro2021csdi}. CSDI leverages conditional score-based diffusion processes for time series modeling and imputation. It generates future or missing values conditioned on observed inputs, offering strong performance in probabilistic forecasting and data completion tasks.

$\bullet$ \textbf{TAD-GNN} (Travel-Time and Availability-aware Diffusion Graph Neural Network)~\cite{Patil2024}. TAD-GNN enhances traffic flow forecasting by incorporating domain-aware graph structures derived from travel-time estimates and data availability scores. It constructs adaptive adjacency matrices using spatial distances and sensor completeness, enabling robust spatial encoding through GCNs and temporal modeling via LSTM.

\subsection{Experiment Settings}
We split the dataset into $80\%$ for training and $20\%$ for testing. All models are trained using the Adam optimizer with automatic mixed precision. Xavier initialization is used to initialize network parameters. Training is conducted on an NVIDIA GeForce RTX 4090 GPU.

We employ tailored training configurations for various modules within our framework to ensure stable optimization and effective convergence. For the CLIP module, we apply a contrastive loss with a temperature of 0.07, using a batch size of 256 and a learning rate of $7 \times 10^{-5}$. We adopt a linear warm-up followed by a cosine learning rate decay schedule. The self-guiding encoder is trained with a learning rate of $5 \times 10^{-5}$ and a cosine decay schedule. For the diffusion model, we use 100 denoising steps with a batch size of 32, following the same initialization and optimizer settings as above. History length and prediction length are set to 24 and 12, respectively. 

{
In terms of modality configuration, we categorize baselines into single-modal and multi-modal settings. Specifically, LSTM, STGCN, and TimeGrad are implemented in a single-modal manner and are provided only with historical private car activity data as input, consistent with their original designs. In contrast, all other baseline models (i.e., DiffSTG, USTD, TACTiS, CSDI, and TAD-GNN) are configured in a multi-modal setting, where both satellite imagery and POI-based features are incorporated as external environmental conditions. The extracted multimodal embeddings were concatenated with the spatiotemporal hidden representations before the final prediction layer for all multimodal baselines, ensuring identical fusion depth and parameter scale across models.
}

\begin{table*}[tb]
\centering
\footnotesize
\vspace{-3mm}
\caption{Results comparison on forecasting across three cities. Smaller values indicate better performance except for $R^2$. Bold indicates the best overall result; underline denotes the second-best within deterministic or probabilistic methods. The reported mean and standard error are obtained from 10 re-training and evaluation independent runs.}
\vspace{-2mm}
\label{tab:baseline}
\begin{tabular}{lccccccc}
\toprule
\multicolumn{8}{c}{\textbf{Changsha}} \\
\midrule
\textbf{Metrics} & \textbf{MAE} & \textbf{RMSE} & \textbf{$R^2$} & \textbf{NMAE} & \textbf{NRMSE} & \textbf{CRPS} & \textbf{CE} \\
\midrule
LSTM & 0.9148{\scriptsize$\pm$0.0286} & 1.5947{\scriptsize$\pm$0.0179} & 0.9497{\scriptsize$\pm$0.0170} & 0.1532{\scriptsize$\pm$0.0062} & 0.0575{\scriptsize$\pm$0.0204} & -- & --\\
STGCN & 0.5922{\scriptsize$\pm$0.0117} & 1.4492{\scriptsize$\pm$0.0075} & 0.9575{\scriptsize$\pm$0.0173} & 0.1527{\scriptsize$\pm$0.0111} & 0.0473{\scriptsize$\pm$0.0052} & -- & -- \\
TimeGrad & 0.6058{\scriptsize$\pm$0.0164} & 1.3593{\scriptsize$\pm$0.0054} & 0.9273{\scriptsize$\pm$0.0070} & 0.1463{\scriptsize$\pm$0.0042} & 0.0454{\scriptsize$\pm$0.0041} & 0.2231{\scriptsize$\pm$0.0091} & 0.0935{\scriptsize$\pm$0.0021} \\
\midrule
USTD & 0.5773{\scriptsize$\pm$0.0041} & 1.3087{\scriptsize$\pm$0.0038} & 0.9586{\scriptsize$\pm$0.0085} & 0.1378{\scriptsize$\pm$0.0022} & 0.0432{\scriptsize$\pm$0.0017} & 0.2326{\scriptsize$\pm$0.0067} & 0.0793{\scriptsize$\pm$0.0019}\\
DiffSTG & 0.5882{\scriptsize$\pm$0.0062} & 1.3067{\scriptsize$\pm$0.0023} & 0.9578{\scriptsize$\pm$0.0049} & \underline{0.1367{\scriptsize$\pm$0.0015}} & 0.0431{\scriptsize$\pm$0.0032} & \underline{0.2182{\scriptsize$\pm$0.0047}} &\underline{0.0684{\scriptsize$\pm$0.0016}} \\
TACTiS & \underline{0.5601{\scriptsize$\pm$0.0117}} & \underline{1.2453{\scriptsize$\pm$0.0029}} & \underline{0.9615{\scriptsize$\pm$0.0055}} & 0.1374{\scriptsize$\pm$0.0038} & \underline{0.0430{\scriptsize$\pm$0.0023}} & 0.2223{\scriptsize$\pm$0.0068} & 0.0712{\scriptsize$\pm$0.0018}\\
CSDI & 0.5805{\scriptsize$\pm$0.0077} & 1.3193{\scriptsize$\pm$0.0063} & 0.9305{\scriptsize$\pm$0.0071} & 0.1410{\scriptsize$\pm$0.0049} & 0.0442{\scriptsize$\pm$0.0034} & 0.2512{\scriptsize$\pm$0.0049} & 0.0965{\scriptsize$\pm$0.0023}\\
TAD-GNN & 0.5833{\scriptsize$\pm$0.0051} & 1.3222{\scriptsize$\pm$0.0052} & 0.9300{\scriptsize$\pm$0.0048} & 0.1798{\scriptsize$\pm$0.0041} & 0.0499{\scriptsize$\pm$0.0039} & 0.2302{\scriptsize$\pm$0.0050} & 0.0898{\scriptsize$\pm$0.0022}\\
\rowcolor{gray!20} \textbf{MAEDE} & \textbf{0.5368{\scriptsize$\pm$0.0094}} & \textbf{1.1874{\scriptsize$\pm$0.0057}} & \textbf{0.9652{\scriptsize$\pm$0.0043}} & \textbf{0.1304{\scriptsize$\pm$0.0025}} & \textbf{0.0409{\scriptsize$\pm$0.0039}} & \textbf{0.1836{\scriptsize$\pm$0.0053}} & \textbf{0.0572{\scriptsize$\pm$0.0014}}\\
\rowcolor{gray!5} \textit{Improvement} & $4.16\%$ & $4.64\%$ & $0.38\%$ & $5.09\%$ & $4.88\%$ & $17.42\%$ & $16.38\%$ \\
\midrule
\multicolumn{8}{c}{\textbf{Guangzhou}} \\
\midrule
\textbf{Metrics} & \textbf{MAE} & \textbf{RMSE} & \textbf{$R^2$} & \textbf{NMAE} & \textbf{NRMSE} & \textbf{CRPS} & \textbf{CE} \\
\midrule
LSTM & 0.1420{\scriptsize$\pm$0.0082} & 1.6399{\scriptsize$\pm$0.0154} & 0.9483{\scriptsize$\pm$0.0092} & 0.1500{\scriptsize$\pm$0.0053} & 0.0584{\scriptsize$\pm$0.0124} & -- & -- \\
STGCN & 0.1525{\scriptsize$\pm$0.0093} & 1.5165{\scriptsize$\pm$0.0061} & 0.9511{\scriptsize$\pm$0.0072} & 0.1574{\scriptsize$\pm$0.0048} & 0.0487{\scriptsize$\pm$0.0036} & -- & -- \\
TimeGrad & 0.1461{\scriptsize$\pm$0.0077} & 1.3858{\scriptsize$\pm$0.0079} & 0.9268{\scriptsize$\pm$0.0062} & 0.1496{\scriptsize$\pm$0.0031} & 0.0467{\scriptsize$\pm$0.0041} & 0.2239{\scriptsize$\pm$0.0061} & 0.0948{\scriptsize$\pm$0.0020}\\
\midrule
USTD & 0.1378{\scriptsize$\pm$0.0044} & \underline{1.3193{\scriptsize$\pm$0.0052}} & 0.9576{\scriptsize$\pm$0.0046} & 0.1400{\scriptsize$\pm$0.0037} & \underline{0.0438{\scriptsize$\pm$0.0022}} & \underline{0.2190{\scriptsize$\pm$0.0038}} & \underline{0.0709{\scriptsize$\pm$0.0017}} \\
DiffSTG & 0.1361{\scriptsize$\pm$0.0033} & 1.3204{\scriptsize$\pm$0.0043} & 0.9574{\scriptsize$\pm$0.0055} & 0.1386{\scriptsize$\pm$0.0029} & 0.0455{\scriptsize$\pm$0.0015} & 0.2347{\scriptsize$\pm$0.0045} & 0.0783{\scriptsize$\pm$0.0021} \\
TACTiS & \underline{0.1358{\scriptsize$\pm$0.0048}} & 1.3208{\scriptsize$\pm$0.0060} & \underline{0.9582{\scriptsize$\pm$0.0039}} & \underline{0.1368{\scriptsize$\pm$0.0035}} & 0.0453{\scriptsize$\pm$0.0024} & 0.2217{\scriptsize$\pm$0.0063} & 0.0721{\scriptsize$\pm$0.0019}\\
CSDI & 0.1387{\scriptsize$\pm$0.0022} & 1.3347{\scriptsize$\pm$0.0031} & 0.9514{\scriptsize$\pm$0.0039} & 0.1452{\scriptsize$\pm$0.0017} & 0.0472{\scriptsize$\pm$0.0020} & 0.2354{\scriptsize$\pm$0.0032} & 0.0826{\scriptsize$\pm$0.0021} \\
TAD-GNN & 0.1384{\scriptsize$\pm$0.0036} & 1.3342{\scriptsize$\pm$0.0025} & 0.9513{\scriptsize$\pm$0.0041} & 0.1385{\scriptsize$\pm$0.0015} & 0.0487{\scriptsize$\pm$0.0028} & 0.2510{\scriptsize$\pm$0.0014} & 0.0853{\scriptsize$\pm$0.0016} \\
\rowcolor{gray!20} \textbf{MAEDE} & \textbf{0.1304{\scriptsize$\pm$0.0047}} & \textbf{1.2103{\scriptsize$\pm$0.0053}} & \textbf{0.9614{\scriptsize$\pm$0.0052}} & \textbf{0.1259{\scriptsize$\pm$0.0043}} & \textbf{0.0429{\scriptsize$\pm$0.0038}} & \textbf{0.1895{\scriptsize$\pm$0.0047}} & \textbf{0.0606{\scriptsize$\pm$0.0015}} \\
\rowcolor{gray!5} \textit{Improvement} & $3.97\%$ & $8.26\%$ & $0.33\%$ & $7.98\%$ & $2.05\%$ & $13.48\%$ & $14.52\%$ \\
\midrule
\multicolumn{8}{c}{\textbf{Shenzhen}} \\
\midrule
\textbf{Metrics} & \textbf{MAE} & \textbf{RMSE} & \textbf{$R^2$} & \textbf{NMAE} & \textbf{NRMSE} & \textbf{CRPS} & \textbf{CE} \\
\midrule
LSTM & 1.6392{\scriptsize$\pm$0.0113} & 1.6520{\scriptsize$\pm$0.0105} & 0.9421{\scriptsize$\pm$0.0081} & 0.1522{\scriptsize$\pm$0.0031} & 0.0582{\scriptsize$\pm$0.0028} & -- & -- \\
STGCN & 1.5164{\scriptsize$\pm$0.0083} & 1.5267{\scriptsize$\pm$0.0072} & 0.9494{\scriptsize$\pm$0.0044} & 0.1591{\scriptsize$\pm$0.0029} & 0.0494{\scriptsize$\pm$0.0022} & -- & -- \\
TimeGrad & 1.3858{\scriptsize$\pm$0.0091} & 1.4566{\scriptsize$\pm$0.0095} & 0.9264{\scriptsize$\pm$0.0056} & 0.1519{\scriptsize$\pm$0.0035} & 0.0472{\scriptsize$\pm$0.0033} & 0.2251{\scriptsize$\pm$0.0048} & 0.0938{\scriptsize$\pm$0.0020}\\
\midrule
USTD & 1.3199{\scriptsize$\pm$0.0050} & 1.3328{\scriptsize$\pm$0.0036} & \underline{0.9524{\scriptsize$\pm$0.0032}} & 0.1412{\scriptsize$\pm$0.0036} & \underline{0.0447{\scriptsize$\pm$0.0020}} & 0.2358{\scriptsize$\pm$0.0034} & \underline{0.0728{\scriptsize$\pm$0.0017}}\\
DiffSTG & 1.3203{\scriptsize$\pm$0.0031} & \underline{1.3293{\scriptsize$\pm$0.0026}} & 0.9500{\scriptsize$\pm$0.0037} & 0.1401{\scriptsize$\pm$0.0024} & 0.0453{\scriptsize$\pm$0.0018} & \underline{0.2271{\scriptsize$\pm$0.0044}} & 0.0745{\scriptsize$\pm$0.0018}\\
TACTiS & \underline{1.2889{\scriptsize$\pm$0.0042}} & 1.3308{\scriptsize$\pm$0.0040} & 0.9421{\scriptsize$\pm$0.0028} & \underline{0.1364{\scriptsize$\pm$0.0019}} & 0.0462{\scriptsize$\pm$0.0025} & 0.2310{\scriptsize$\pm$0.0026} & 0.0782{\scriptsize$\pm$0.0019}\\
CSDI & 1.3421{\scriptsize$\pm$0.0029} & 1.3592{\scriptsize$\pm$0.0030} & 0.9382{\scriptsize$\pm$0.0026} & 0.1503{\scriptsize$\pm$0.0015} & 0.0469{\scriptsize$\pm$0.0016} & 0.2367{\scriptsize$\pm$0.0023} & 0.0847{\scriptsize$\pm$0.0016}\\
TAD-GNN & 1.3423{\scriptsize$\pm$0.0047} & 1.3585{\scriptsize$\pm$0.0045} & 0.9381{\scriptsize$\pm$0.0038} & 0.1422{\scriptsize$\pm$0.0024} & 0.0459{\scriptsize$\pm$0.0027} & 0.2380{\scriptsize$\pm$0.0032} & 0.0813{\scriptsize$\pm$0.0015}\\
\rowcolor{gray!20} \textbf{MAEDE} & \textbf{1.2103{\scriptsize$\pm$0.0031}} & \textbf{1.2298{\scriptsize$\pm$0.0037}} & \textbf{0.9525{\scriptsize$\pm$0.0020}} & \textbf{0.1246{\scriptsize$\pm$0.0016}} & \textbf{0.0420{\scriptsize$\pm$0.0020}} & \textbf{0.1967{\scriptsize$\pm$0.0024}} & \textbf{0.0626{\scriptsize$\pm$0.0014}}\\
\rowcolor{gray!5} \textit{Improvement} & $6.09\%$ & $7.49\%$ & $0.01\%$ & $8.67\%$ & $6.04\%$ & $13.40\%$ & $13.97\%$\\
\bottomrule
\end{tabular}
\vspace{-3mm}
\end{table*}

\subsection{Metrics}
We utilize Mean Absolute Error (MAE), Mean Squared Error (MSE), Root Mean Squared Error (RMSE), Normalized Root Mean Squared Error (NRMSE), Normalized Mean Absolute Error (NMAE), and Coefficient of Determination ($R^2$) as evaluation metrics.
For probabilistic models, we additionally report the Continuous Ranked Probability Score (CRPS), Calibration Error (CE), CRPS\(_\text{sum}\), NMAE\(_\text{sum}\), and NRMSE\(_\text{sum}\) by sampling 100 predictions. \textcolor{black}{CRPS is a univariate strictly proper scoring rule which measures the compatibility of a cumulative distribution function $F$ with an observation $x \in \mathbb{R}$ as:}
\begin{equation}
    \textcolor{black}{\mathrm{CRPS}(F, x) = \int_{\mathbb{R}} \big(F(y) - \mathbf{I}(x \le y)\big)^2 \, dy,}
\end{equation}
\textcolor{black}{where $I(x\le y)$ is the indicator function, which is 1 if $x \le y$ and 0 otherwise. The CRPS attains the minimum value when the predictive distribution $F$ same as the data distribution. Deterministic metrics evaluate point-wise prediction accuracy by comparing a single predicted value against the observation, thus reflecting the model’s average error but not its uncertainty representation. In contrast, probabilistic metrics assess the quality of the entire predictive distribution. CRPS evaluates both the sharpness and the accuracy of the predicted distribution, while calibration error measures the consistency between predicted uncertainty and empirical frequencies. These metrics reveal whether a model is overconfident, underconfident, or well-calibrated, which cannot be captured by deterministic errors alone. This complementary metric design allows us to assess both prediction accuracy and uncertainty-aware behavior, which is particularly important for diffusion-based stochastic forecasting models.}

\begin{figure}[tb]
\centering
\vspace{-2mm}
\subfigure[\textcolor{black}{Forecasting calibration plot}]{\includegraphics[width=0.45\linewidth]{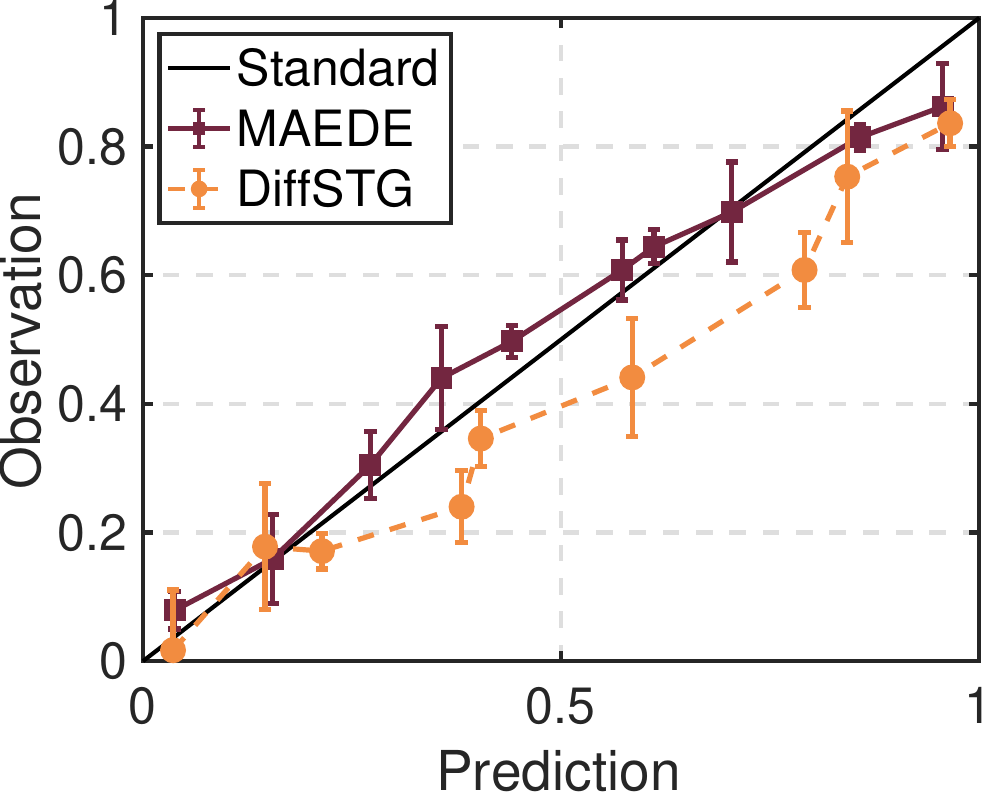}}
\subfigure[\textcolor{black}{Sensitivity to parameter $\rho$}]{\includegraphics[width=0.45\linewidth]{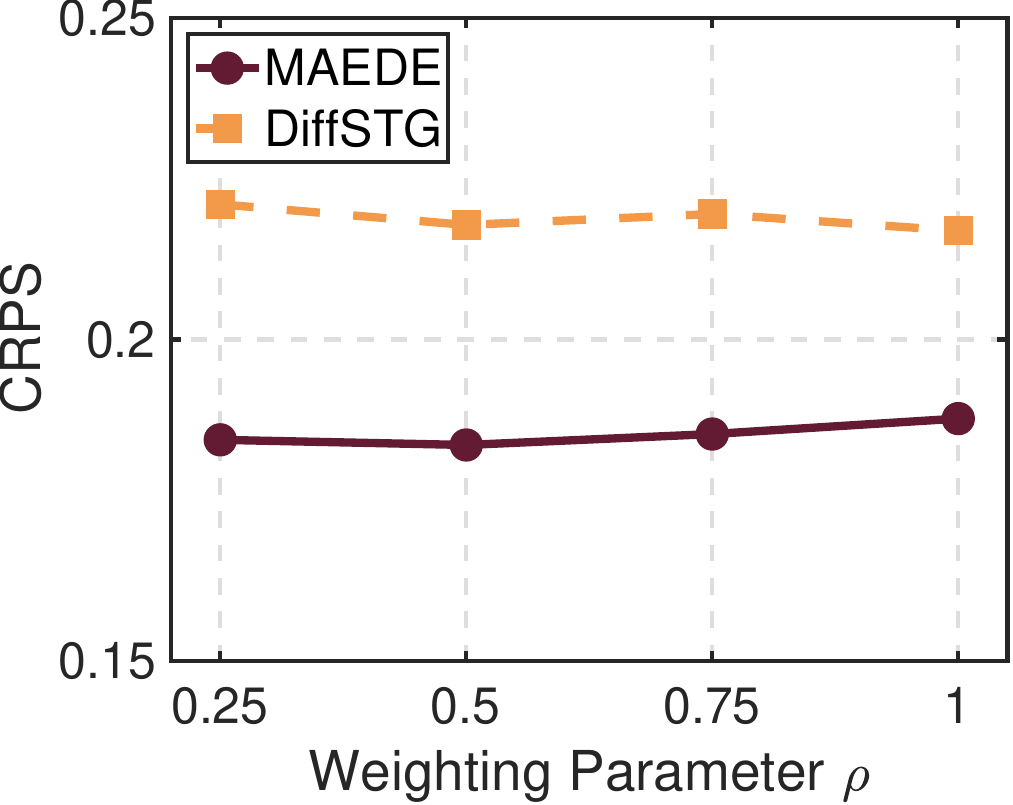}}
\vspace{-2mm}
\caption{\textcolor{black}{(a) MAEDE demonstrates superior calibration, with predicted probabilities closely matching observed frequencies, while DiffSTG exhibits noticeable miscalibration at low probability regions. (b) MAEDE is robust to various choices of weighting parameter $\rho$. }}
\label{fig:sentivity_to_prediction_length_and_rho}
\vspace{-2mm}
\end{figure}


\subsection{Overall Performance Evaluation}

{
The experimental results of MAEDE and baseline methods are summarized in Table~\ref{tab:baseline}.
Across all three cities, MAEDE achieves the best overall performance on every evaluation metric.
Specifically, it reduces the MAE by an average of $4.7\%$, RMSE by $6.8\%$, and CRPS by $14.8\%$ compared with the strongest baseline in each category, demonstrating its superior capability in modeling complex urban dynamics.
}
{
Compared with single-modal baselines, MAEDE achieves lower prediction errors and higher fitting accuracy. For instance, in the Changsha dataset, MAEDE attains a lower MAE (0.5368 vs. 0.6058 for TimeGrad) and a higher $R^2$ (0.9652 vs. 0.9497 for LSTM). This improvement stems from MAEDE’s integration of multi-modal environmental features that encode both static spatial structures and semantic textual cues, enriching the model’s understanding of urban mobility patterns. Relative to other multi-modal probabilistic forecasting baselines, MAEDE also yields notable improvements in accuracy and uncertainty-aware metrics. For example, MAEDE achieves lower RMSE (1.1874 vs. 1.2453 for TACTiS), lower CRPS (0.1836 vs. 0.2182 for DiffSTG), and better calibration error (0.0572 vs. 0.0684 for DiffSTG). These advantages can be attributed to MAEDE’s soft-attention mechanism, which adaptively emphasizes complementary feature contributions to refine multi-granular representations, and its multi-granularity diffusion design, which allows the model to learn long-term mobility trends and short-term fluctuations jointly. Together, these modules enhance temporal coherence, spatial awareness, and calibration reliability across diverse urban settings. 
}
{
Furthermore, MAEDE exhibits notably better probabilistic calibration compared to other baselines. As illustrated in Fig.~\ref{fig:sentivity_to_prediction_length_and_rho}(a), the predicted uncertainty of MAEDE aligns closely with the observed distribution, yielding a lower Calibration Error (CE = 0.0572) than DiffSTG (CE = 0.0684). This indicates that MAEDE not only achieves superior point forecasts but also provides well-calibrated uncertainty estimates, ensuring reliable confidence intervals for informed decision-making in real-world applications.
}
\textcolor{black}{
We also conducted experiments to analyze the affect of the weighting parameter $\rho$. Specifically, we vary $\rho$ from 0.25 to 1.0 and evaluate MAEDE on the Changsha dataset. The results illustrated in Fig.~\ref{fig:sentivity_to_prediction_length_and_rho} show that the performance remains highly stable across different values of $\rho$, indicating that MAEDE is robust to various choices of $\rho$ and stably integrates stationary information and dynamic mobility patterns. Based on this observation, we set $\rho = 0.5$ as the default value in all experiments.}

The superior performance of MAEDE results from the synergistic integration of several key architectural components: 
(1) a language-guided static environmental encoder based on the LLM-enhanced CLIP module, which extracts semantically rich representations of regional functionality; 
(2) a multi-granularity temporal mechanism that captures temporal dependencies across multiple time scales through coarse-to-fine supervision; 
(3) a self-guiding encoder that unifies static and dynamic signals into a coherent spatiotemporal representation; and 
(4) a soft attention fusion mechanism that adaptively emphasizes salient contextual cues across granularities. 

\subsection{Case Study}
\begin{figure*}[tb]  
    \centering
    \includegraphics[width=0.8\textwidth]{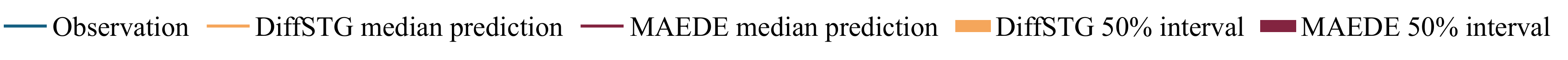}\\
    \vspace{-2mm}
    \subfigure[Office Region]{ \includegraphics[width=0.23\linewidth]{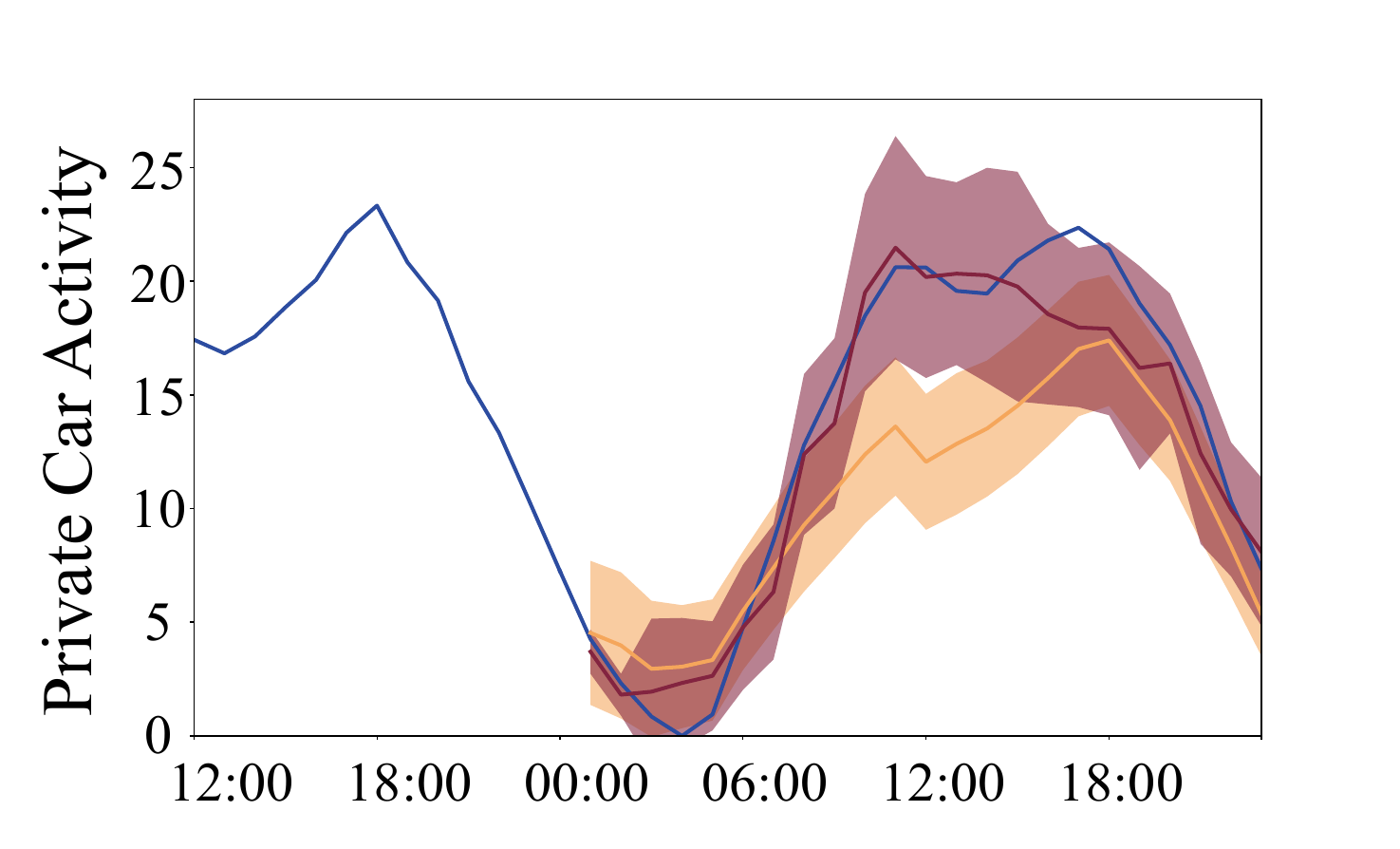}}  
        \subfigure[Residential Region]{ \includegraphics[width=0.23\linewidth]{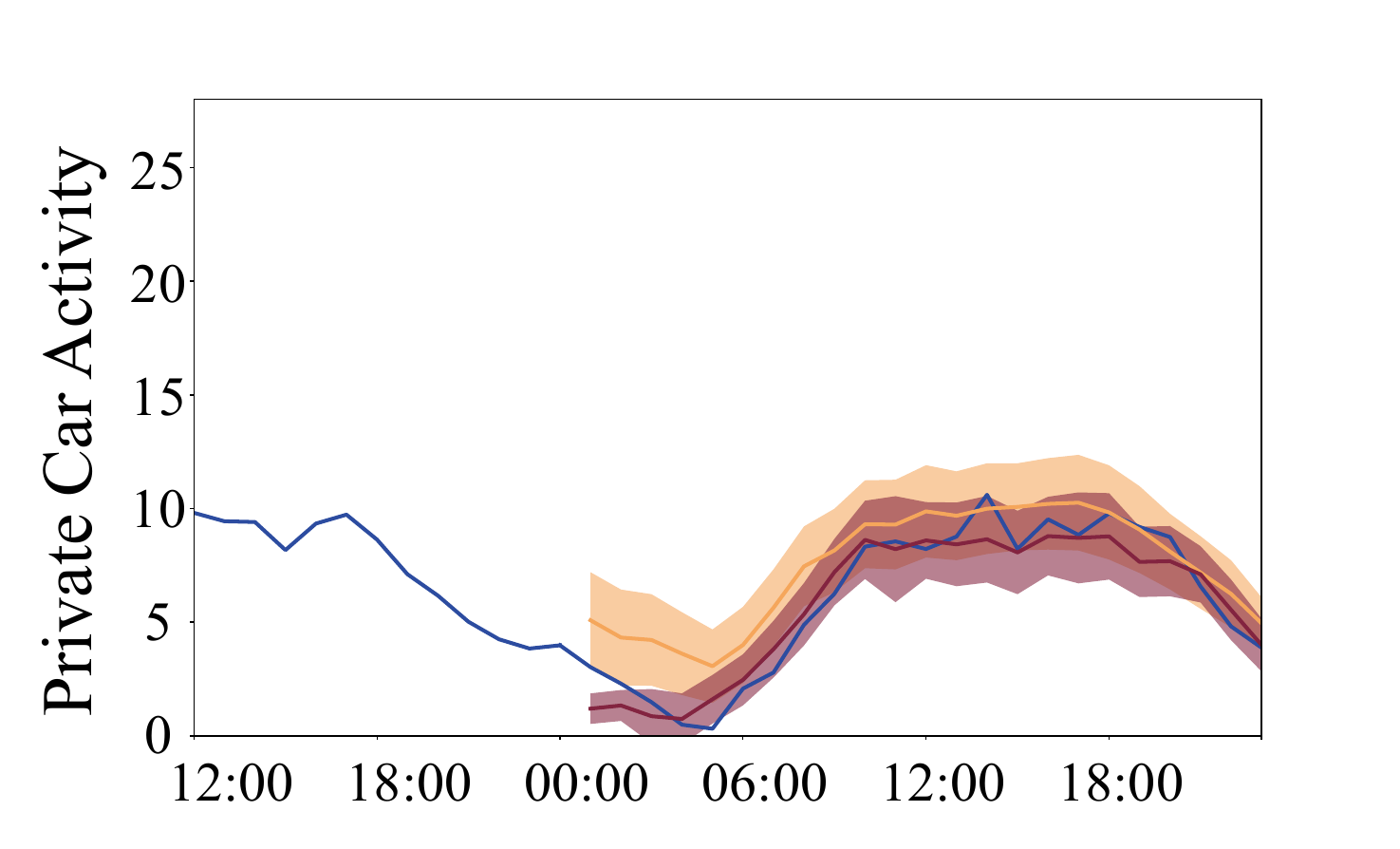}}  
            \subfigure[Entertainment Region]{ \includegraphics[width=0.23\linewidth]{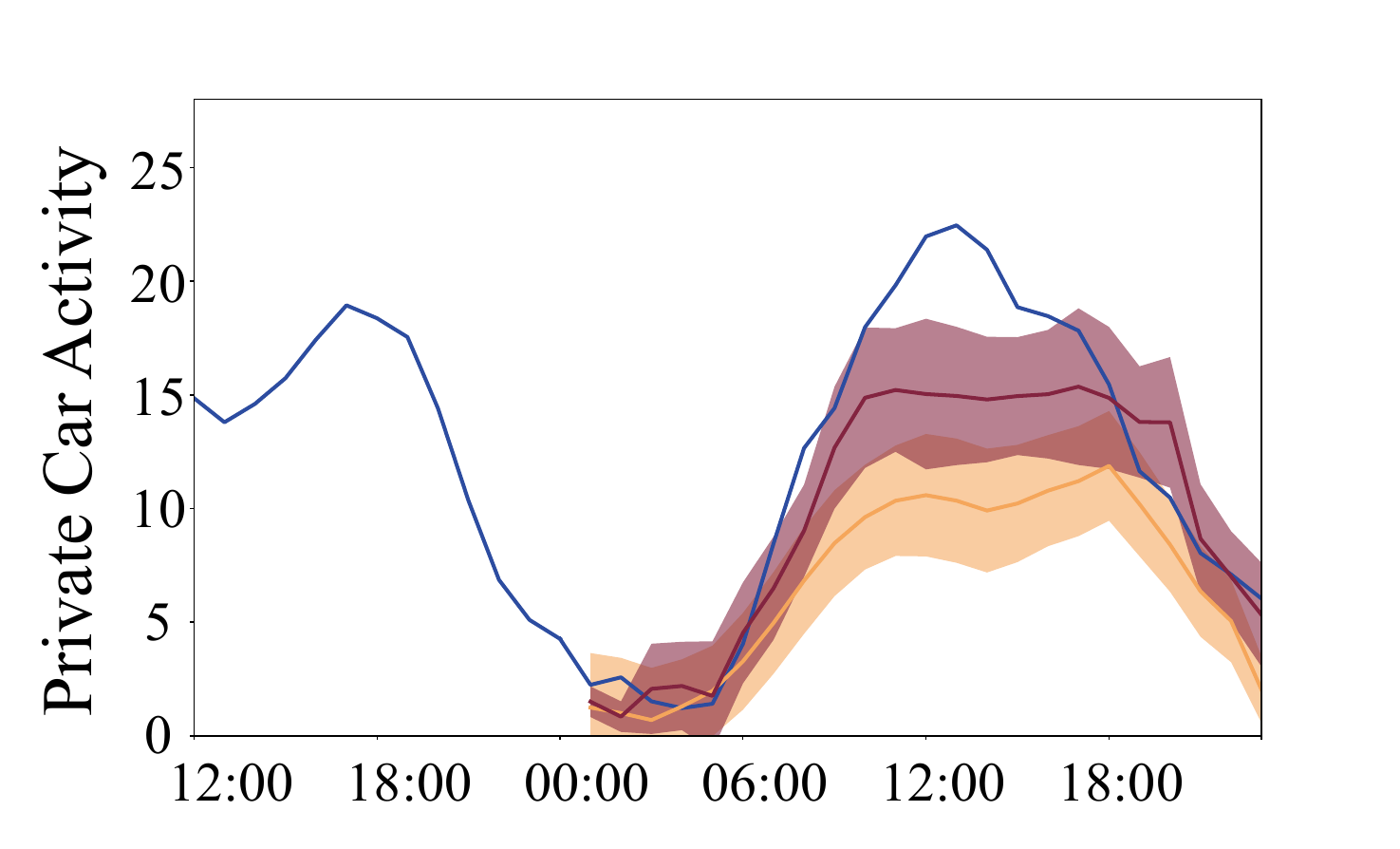}}  
                \subfigure[School Region]{ \includegraphics[width=0.23\linewidth]{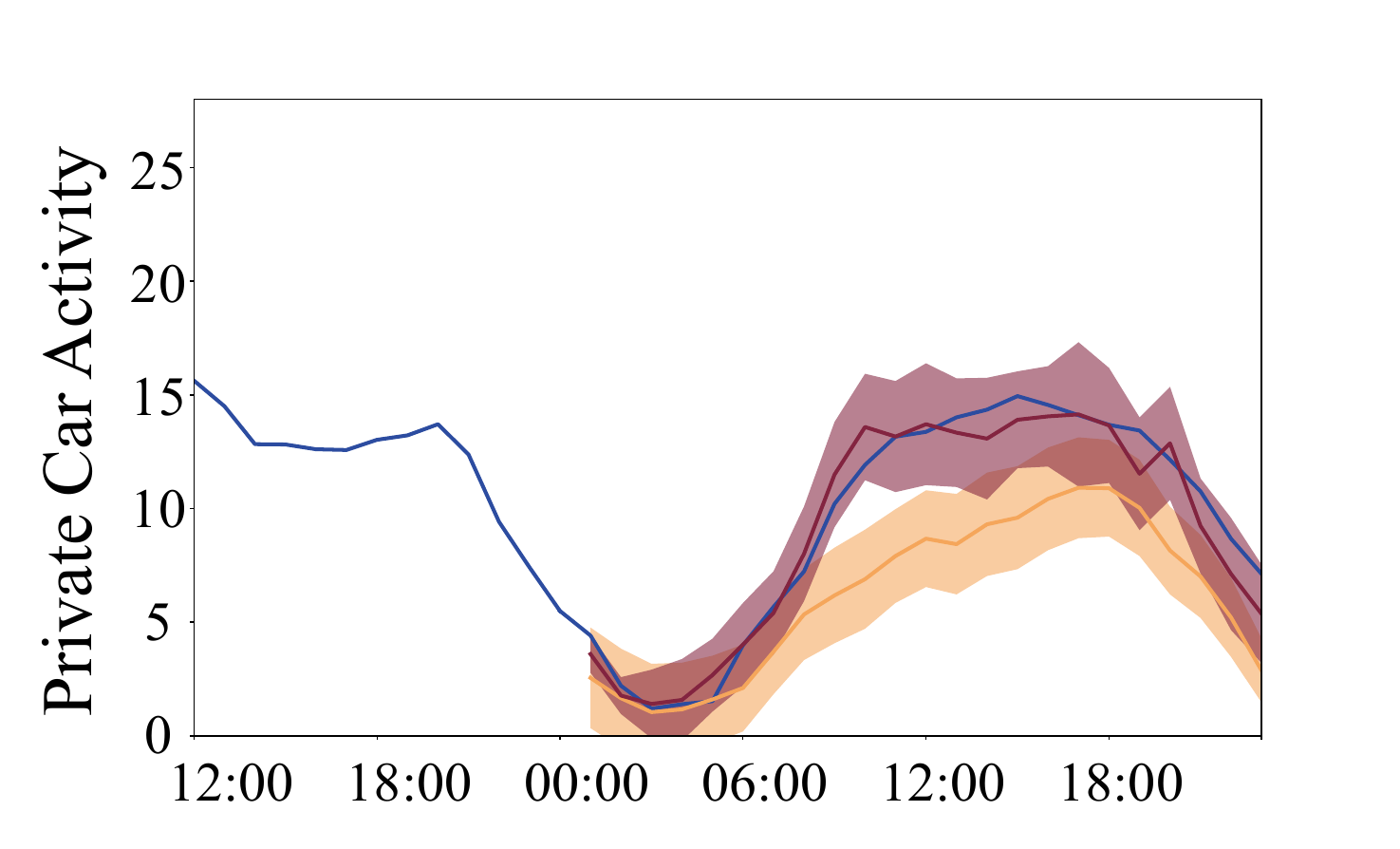}}  
\vspace{-2mm}
    \caption{Case Study on different functional regions. The blue line represents the observed activity, while the median predictions of DiffSTG and MAEDE are shown in orange and red, respectively. Shaded areas indicate the $50\%$ confidence intervals.}
    \label{fig:case study}
\vspace{-2mm}
\end{figure*}

To gain insights into the uncertainty estimates and prediction accuracy, we select four representative regions in Changsha and conduct case studies comparing the performance of MAEDE with the baseline model. Fig.~\ref{fig:case study} illustrates the prediction of $T'=24$ future time steps based on $T=24$ historical time steps. The results indicate that MAEDE achieves superior numerical accuracy compared to DiffSTG, as its predictions align closely with the ground truth. Regarding uncertainty estimation, MAEDE demonstrates higher reliability because the shadow straps encompass the true series with a narrower sampling range. MAEDE can extract high-quality spatio-temporal dependencies from historical data, benefiting from the multi-modal features and multi-granularity self-guidances. 
{
Beyond quantitative performance, these regional forecasts offer actionable insights for urban mobility planning. For instance, the distinct temporal patterns observed across office, residential, entertainment, and school regions can inform dynamic parking allocation, congestion mitigation, and functional zoning adjustments. Planners can leverage the predicted Arrive–Stay–Leave dynamics to identify high-demand periods, optimize parking resources, and adjust traffic control strategies according to regional activity rhythms.
}

\subsection{Ablation Studies}
{
To assess the contribution of each component in MAEDE, we conduct an ablation study by selectively modifying individual modules while keeping the rest of the architecture unchanged. The experiments are performed on the private car trip dataset extracted from Changsha, with both history length and prediction length set to 24 h. The study compares MAEDE with several model variants:
}

$\bullet$ \textbf{w/o Self-Guiding Encoder}: The coarse-grained data is directly fed into the multi-granularity diffusion model, without the self-guiding encoder extracting any additional feature.

$\bullet$ \textbf{w/o Multi-Granularity Diffusion}: Extracted environmental features are simply combined with raw data without explicit modeling of multi-granularity temporal dependencies, and then integrated into USTD for probabilistic forecasting.

$\bullet$ \textbf{w/o LLM-enhanced CLIP}: Multi-granular spatiotemporal dependencies are encoded without static environments.

$\bullet$ \textbf{w/o Soft Attention}: The static environmental features and multi-granular data are processed through the self-guiding encoder, and the output is fed into the multi-granularity Diffusion Model without feature fusion via attention.

$\bullet$ \textcolor{black}{\textbf{w/o POI-feature}: POI information is removed from the LLM-generated textual descriptions, so that the CLIP encoder is enhanced only by generic environmental semantics without explicit POI-based augmentation.}

$\bullet$ \textcolor{black}{\textbf{w/o LLM-generated-text}: the CLIP module uses manually designed template-based text prompts that incorporate POI information, instead of LLM-generated text, for contrastive learning.}

\begin{figure}[tb]
    \centering
    \vspace{-2mm}
    \subfigure{\includegraphics[width=0.7\linewidth]{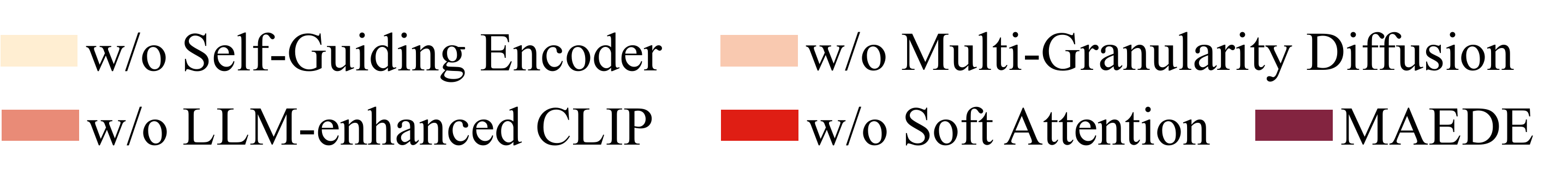}}\\
    \vspace{-2mm}
    \subfigure{\includegraphics[width=0.45\linewidth]{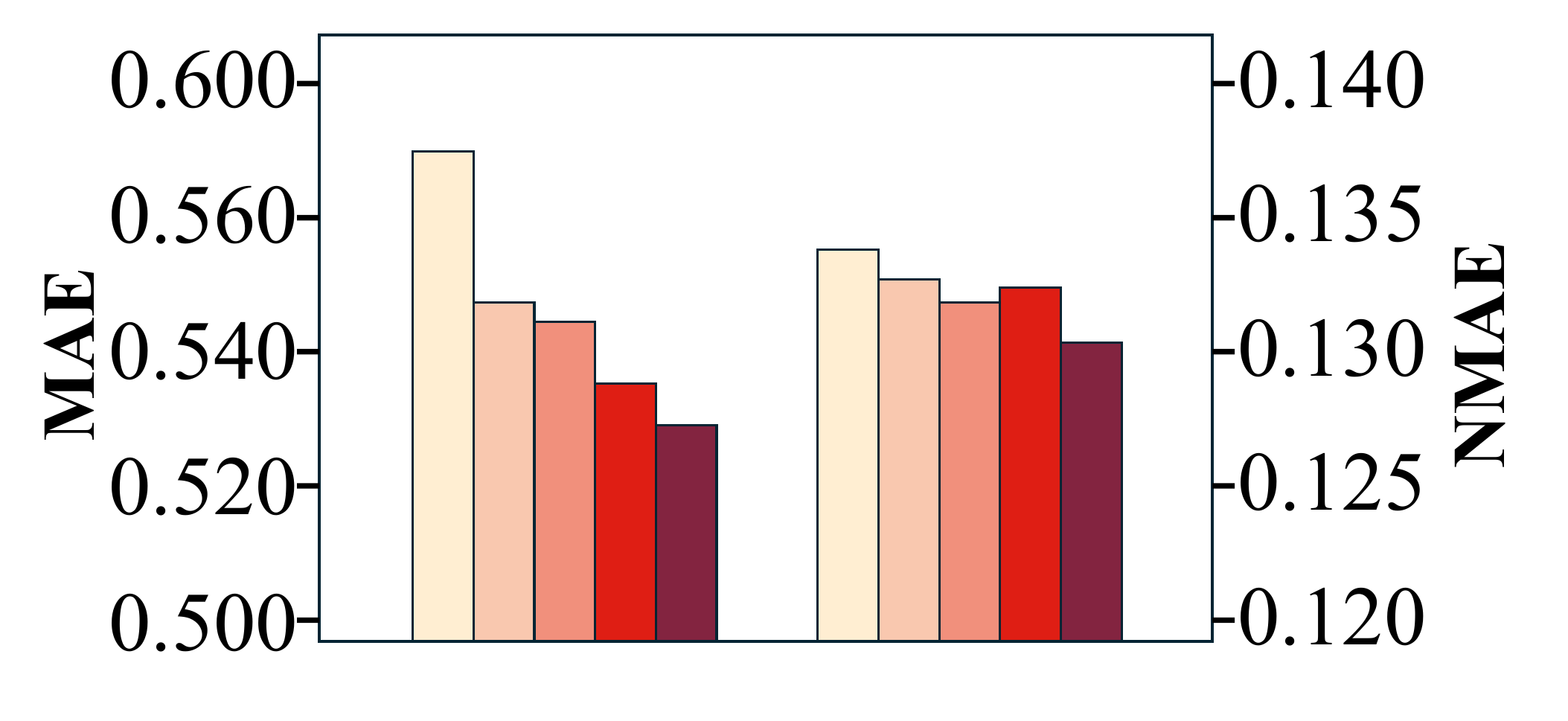}}
    \subfigure{\includegraphics[width=0.45\linewidth]{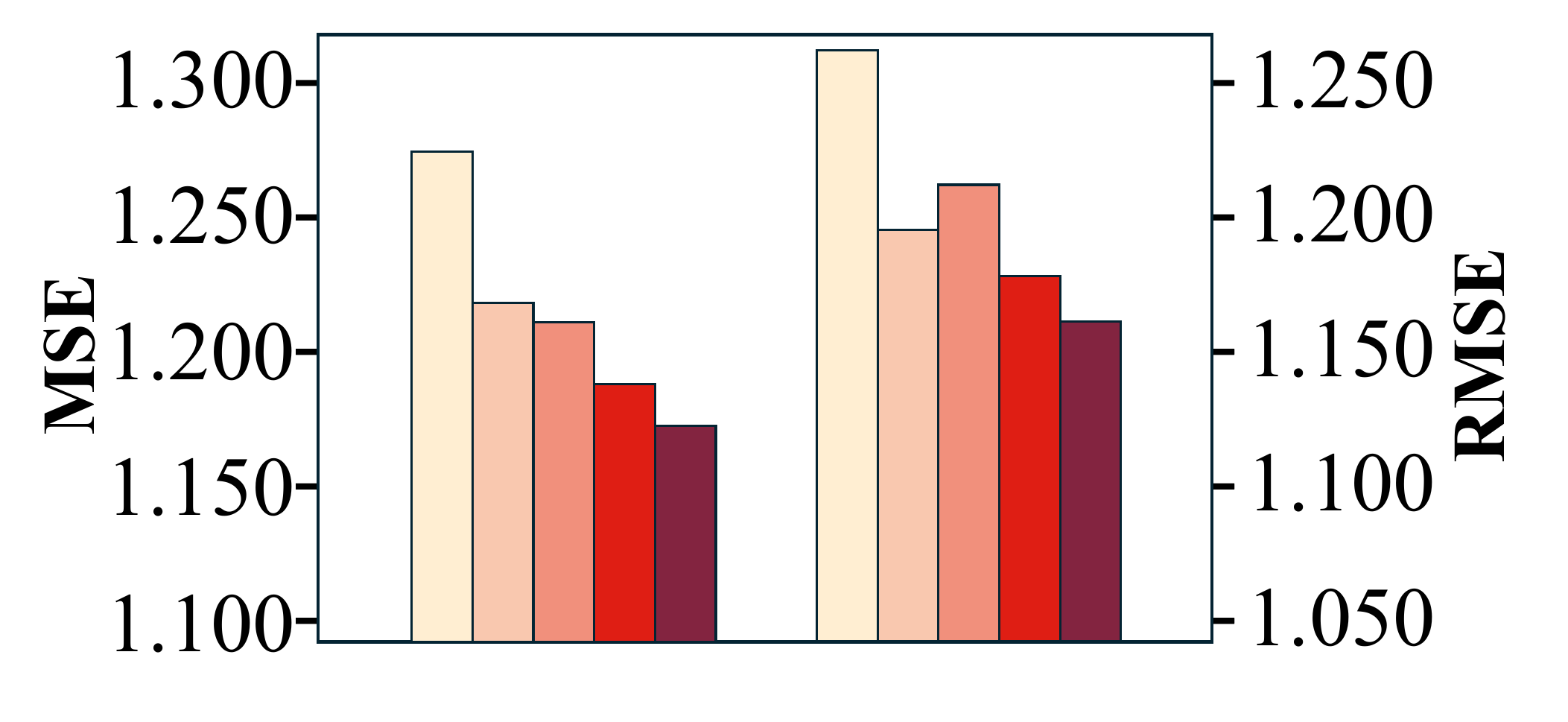}}\\
\vspace{-2mm}
    \subfigure{\includegraphics[width=0.45\linewidth]{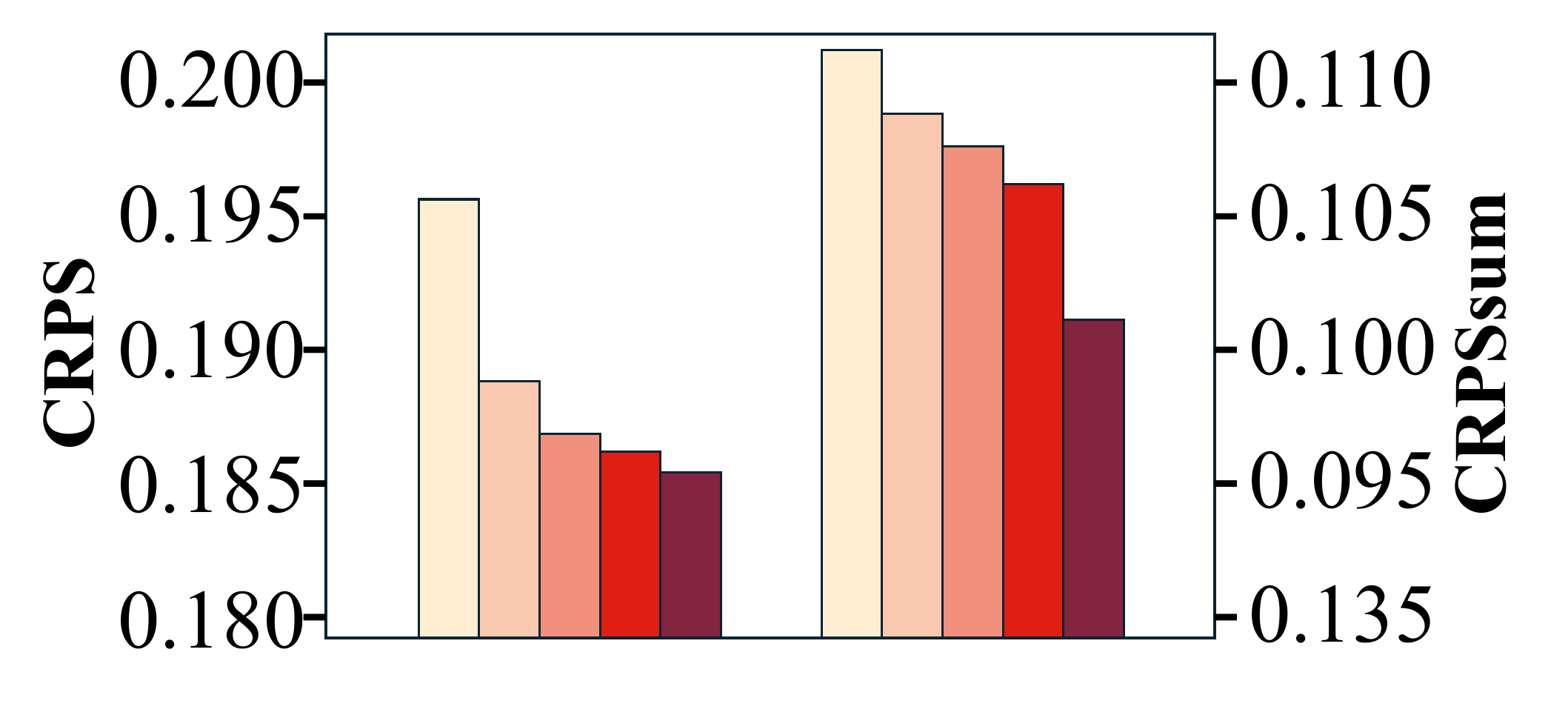}}
    \subfigure{\includegraphics[width=0.45\linewidth]{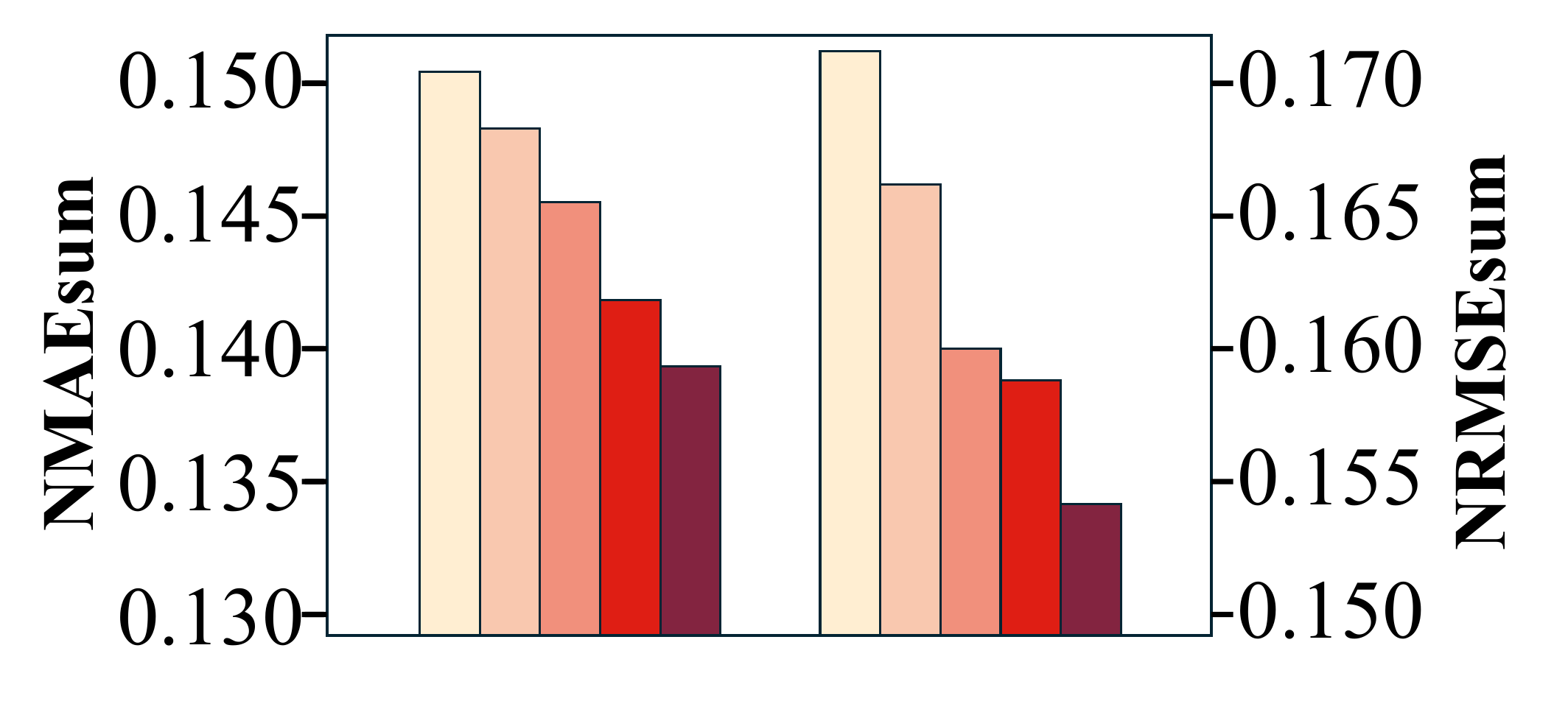}}
    \vspace{-2mm}
    \caption{Ablation study of comparing MAEDE with four variants. Lower indicators denote better performance, and changes in indicators reflect the effect of components.}
    \label{fig:ablation study}
    \vspace{-2mm}
\end{figure}

Fig.~\ref{fig:ablation study} presents the results of an ablation study designed to quantify the impact of each architectural component on forecasting performance. The w/o Self-Guiding Encoder variant exhibits the weakest performance across all metrics, confirming that temporal diffusion alone, without dynamic-static alignment, cannot effectively capture spatiotemporal dependencies. Removing the Multi-Granularity Diffusion module also leads to noticeable degradation, emphasizing the necessity of hierarchical temporal modeling for long-horizon forecasting. Similarly, excluding the LLM-enhanced CLIP reduces performance, indicating that environmental semantics are essential for improving contextual understanding.

{
To further examine the contribution of each module under varying forecasting horizons and across different urban environments, we extend the ablation study to $T = 3, 12,$ and $24$, as summarized in Table~\ref{tab:ablation_length_city}. The results show that each architectural component consistently contributes to the overall performance of MAEDE in both Changsha and Guangzhou. As the prediction length increases, all ablated variants exhibit a more pronounced degradation in MAE and CRPS. In contrast, MAEDE maintains stable performance with only marginal increases in error, demonstrating strong resilience to extended prediction horizons.}\textcolor{black}{ In addition, the ablations w/o POI-feature and w/o LLM-generated-text consistently underperform the full model across all horizons, indicating that both POI-aware semantic enrichment and LLM-generated textual representations play complementary roles in enhancing long-horizon forecasting accuracy.}
Notably, this stability trend holds across both cities, suggesting that MAEDE generalizes effectively to different spatial environments and mobility dynamics. The model’s integrated design enables it to preserve temporal coherence and spatial adaptability.

\subsection{Hyperparameter Study}
\subsubsection{Sensitivity of Granularity}
To assess the impact of multi-granularity configurations on performance, we conduct experiments by varying the multi-granularity dictionary while keeping other hyperparameters fixed. We set both history length and prediction length to 24h, and the model performance is assessed by CRPSsum, NMAEsum, and NRMSEsum. Results is presented in Table~\ref{tab:multi_granularity}, where mg\_dict = 1\_2\_4 corresponds to granularities of 1-hour, 2-hour, and 4-hour intervals.
{
The results demonstrate that incorporating multiple granularities enhances predictive performance compared to single-granularity models. Models with three-granularity configurations consistently outperform those with less granularity, emphasizing the multi-scale temporal representations' advantage. The improvement can be attributed to the model's ability to capture both short-term fluctuations and long-term dependencies, enabling a more comprehensive understanding of temporal structures. However, introducing four granularities does not yield further benefits and leads to slight performance degradation. This phenomenon may be due to an over-distribution of loss weights across multiple granularities, resulting in poor feature learning. Consequently, while multi-granularity modeling is advantageous, excessive granularity could introduce weakly informative patterns, ultimately reducing predictive efficiency.
}

\begin{table}[tb]
    \centering    
    \vspace{-2mm}
    \caption{Performance comparison of different multi-granularity settings. 
    Lower values indicate better predictive accuracy.}
    \label{tab:multi_granularity}
    \vspace{-2mm}
    \footnotesize
    \begin{tabular}{c@{\hspace{1.5mm}}c@{\hspace{1.5mm}}c@{\hspace{1.5mm}}c@{\hspace{1.5mm}}c}
        \toprule
        \textbf{Mg\_Dict} & \textbf{Num\_Gran} & 
        \textbf{CRPSsum} & \textbf{NMAEsum} & \textbf{NRMSEsum} \\
        \midrule
        1       & 1  & 0.2282  & 0.3061  & 0.3369  \\
        1\_2    & 2  & 0.1905  & 0.2617  & 0.3168  \\
        1\_4    & 2  & 0.1904  & 0.2560  & 0.3144  \\
        1\_8    & 2  & 0.1921  & 0.2619  & 0.3101  \\
        1\_2\_4 & 3  & \textbf{0.1723}  & \textbf{0.2223}  & \textbf{0.2841}  \\
        1\_4\_8 & 3  & 0.1846  & 0.2451  & 0.3036  \\
        1\_2\_4\_8 & 4  & 0.1964 & 0.2659 & 0.3242  \\
        \bottomrule
    \end{tabular}
    \vspace{-2mm}
\end{table}

\subsubsection{Sensitivity of Prediction length}

To assess the sensitivity of MAEDE to prediction length and its effectiveness in long-term forecasting, we fix the context length at 24 and extend the prediction horizon to 3, 12, 24, 48, 96, and 144 time steps. A comprehensive evaluation is conducted using six metrics: MAE, NMAE, MSE, RMSE, NRMSE, and CRPS. The results on the Changsha private car trip dataset, as illustrated in Fig.~\ref{fig:sentivity_to_prediction_length_and_rho}, demonstrate that MAEDE maintains strong predictive performance over extended forecasting horizons. { We further conduct 10 independent runs to verify the statistical significance of the results across long-horizon settings. Notably, as the prediction length increases, the performance of MAEDE remains stable without significant degradation. Moreover, MAEDE consistently outperforms the competitive baseline, showing a sustained advantage that suggests its robustness in capturing long-term dependencies. The trend further indicates no signs of convergence between MAEDE and the baseline, reinforcing the model’s superiority in long-term forecasting scenarios.
}

\begin{figure}[tb]
\centering
\subfigure{\includegraphics[width=0.45\linewidth]{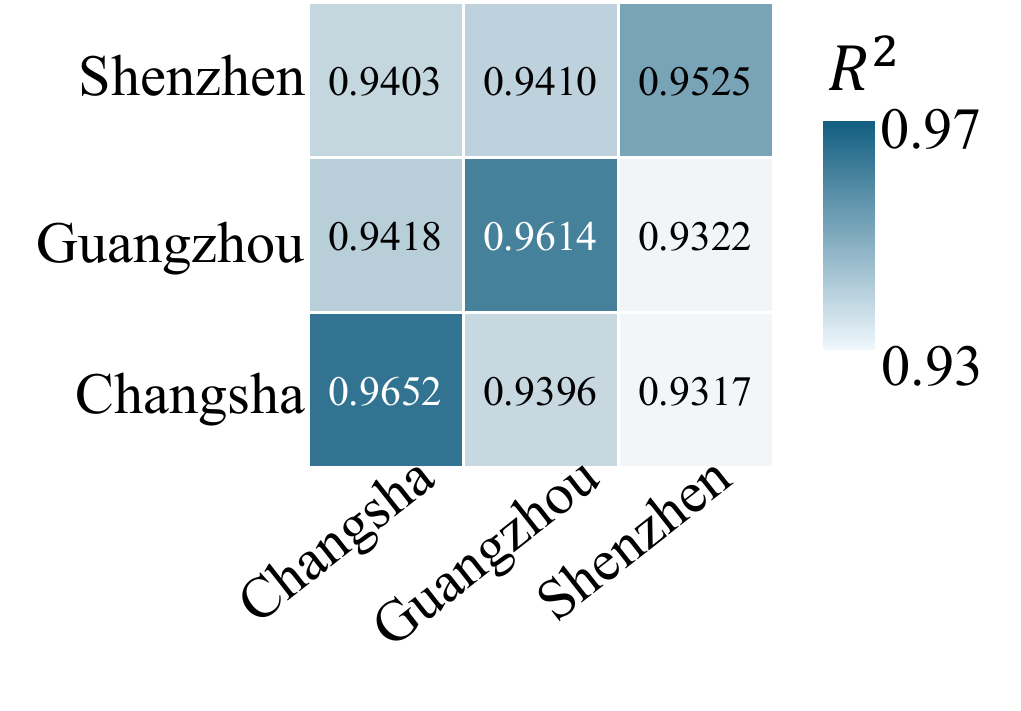}}
\subfigure{\includegraphics[width=0.45\linewidth]{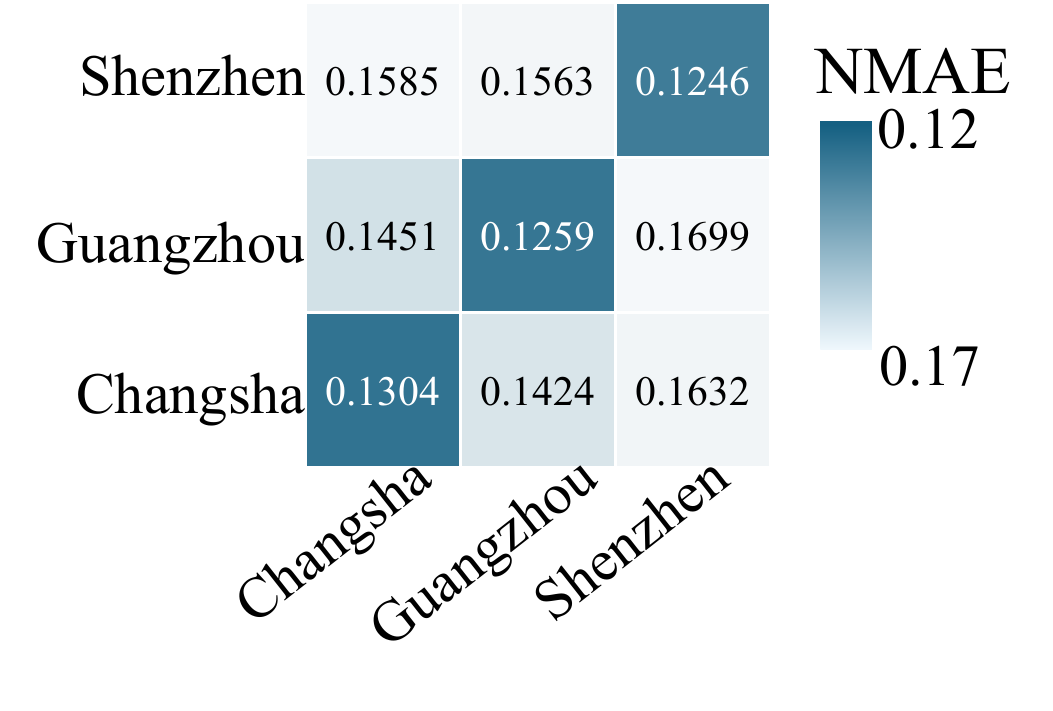}}\\
\subfigure{\includegraphics[width=0.45\linewidth]{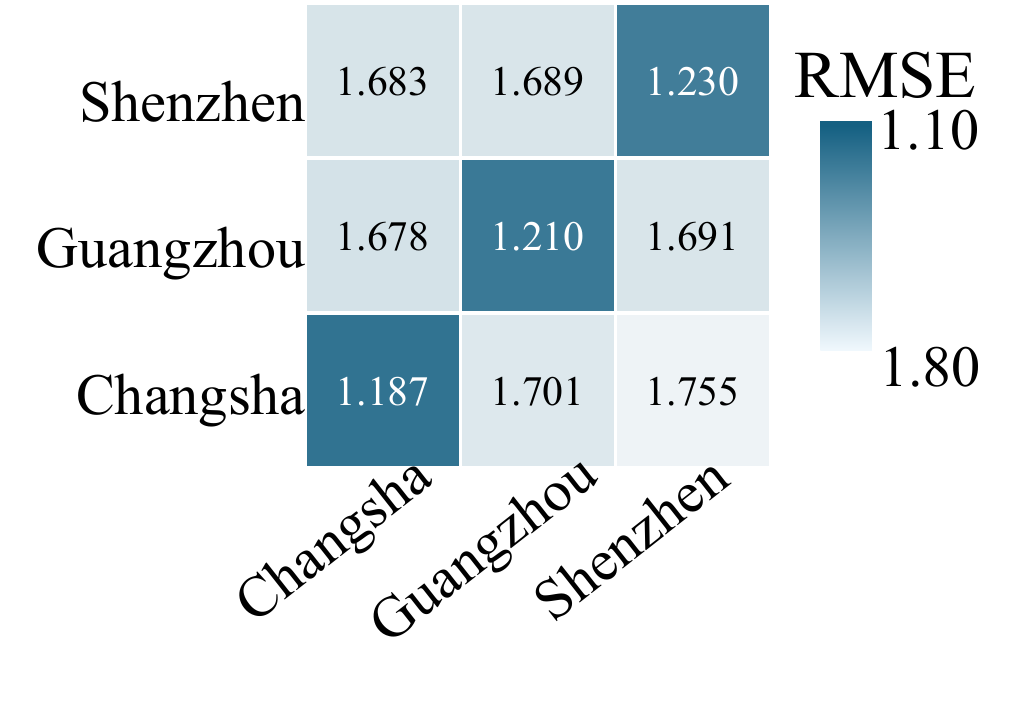}}
\subfigure{\includegraphics[width=0.45\linewidth]{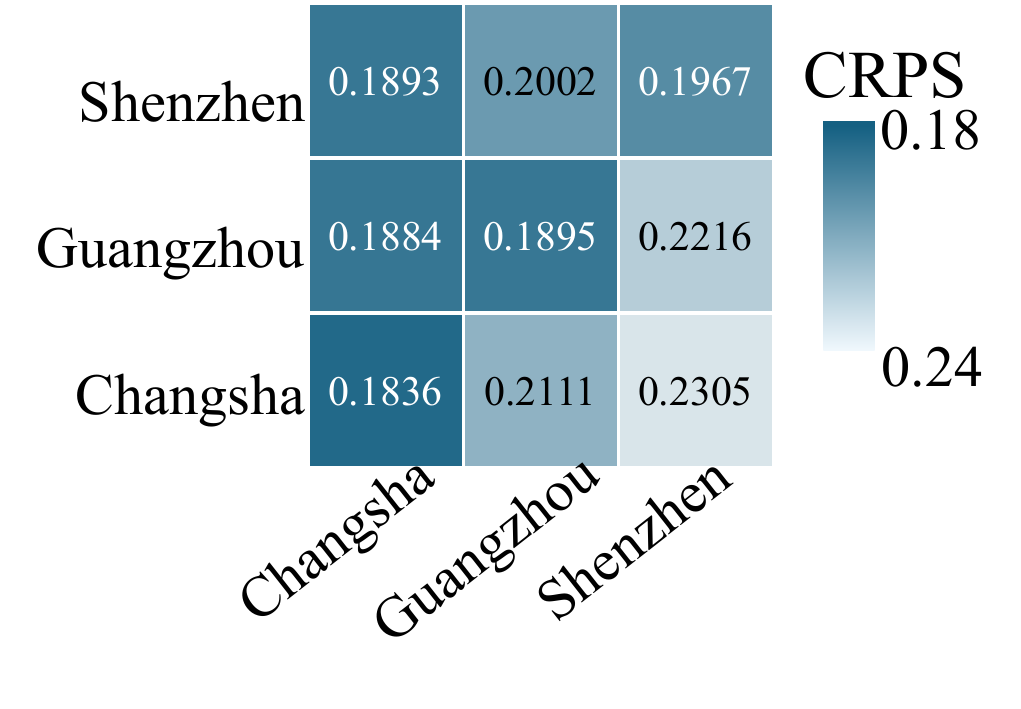}}
\vspace{-2mm}
\caption{Cross-city experiments are conducted on Changsha, Guangzhou, and Shenzhen. Darker grids denote better generalization and robustness.}
\label{fig:Generalization}
\vspace{-2mm}
\end{figure}

\begin{table*}[t]
\centering
\footnotesize
\vspace{-2mm}
\caption{Ablation study of MAEDE under different prediction lengths across Changsh and Guangzhou. Lower values indicate better predictive performance.}
\vspace{-2mm}
\begin{tabular}{lccccccccc}
\toprule
\multirow{2}{*}{\textbf{Ablation Variant}} & 
\multicolumn{3}{c}{\textbf{MAE}} & 
\multicolumn{3}{c}{\textbf{RMSE}} & 
\multicolumn{3}{c}{\textbf{CRPS}} \\
\cmidrule(lr){2-4} \cmidrule(lr){5-7} \cmidrule(lr){8-10}
& $T$=3 & $T$=12 & $T$=24 & $T$=3 & $T$=12 & $T$=24 & $T$=3 & $T$=12 & $T$=24 \\
\midrule
\multicolumn{10}{c}{\textbf{Changsha}} \\
\midrule
w/o Self-Guiding Encoder        & 0.5358 & 0.5664 & 0.6953 & 1.1182 & 1.2581 & 1.6374 & 0.1864 & 0.1968 & 0.2435 \\
w/o LLM-enhanced CLIP           & 0.5237 & 0.5424 & 0.6772 & 1.0755 & 1.2150 & 1.5711 & 0.1795 & 0.1879 & 0.2328 \\
w/o POI-feature                 & 0.5190 & 0.5405 & 0.6738 & 1.0568 & 1.1975 & 1.5532 & 0.1788 & 0.1876 & 0.2246 \\
w/o LLM-generated-text          & 0.5238 & 0.5436 & 0.6769 & 1.0715 & 1.2136 & 1.5719 & 0.1786 & 0.1912 & 0.2304 \\
w/o Multi-Granularity Diffusion & 0.5220 & 0.5481 & 0.6767 & 1.0633 & 1.1981 & 1.5548 & 0.1772 & 0.1868 & 0.2294 \\
w/o Soft Attention              & 0.5142 & 0.5387 & 0.6695 & 1.0487 & 1.1882 & 1.5387 & 0.1784 & 0.1886 & 0.2211 \\
\rowcolor{gray!10}
\textbf{OURS (MAEDE)}           & \textbf{0.5108} & \textbf{0.5368} & \textbf{0.6667} & \textbf{1.0174} & \textbf{1.1303} & \textbf{1.2917} & \textbf{0.1737} & \textbf{0.1836} & \textbf{0.2037} \\
\midrule
\multicolumn{10}{c}{\textbf{Guangzhou}} \\
\midrule
w/o Self-Guiding Encoder        & 0.1431 & 0.1583 & 0.1732 & 1.3014 & 1.3547 & 1.4652 & 0.1928 & 0.1984 & 0.2231 \\
w/o LLM-enhanced CLIP           & 0.1389 & 0.1507 & 0.1664 & 1.2783 & 1.3251 & 1.4215 & 0.1813 & 0.1969 & 0.2115 \\
w/o POI-feature                 & 0.1386 & 0.1496 & 0.1663 & 1.2769 & 1.3260 & 1.4248 & 0.1819 & 0.1966 & 0.2108 \\
w/o LLM-generated-text          & 0.1386 & 0.1505 & 0.1656 & 1.2754 & 1.3158 & 1.4271 & 0.1807 & 0.1965 & 0.2106 \\
w/o Multi-Granularity Diffusion & 0.1368 & 0.1479 & 0.1647 & 1.2634 & 1.3102 & 1.3984 & 0.1805 & 0.1962 & 0.2102 \\
w/o Soft Attention              & 0.1347 & 0.1453 & 0.1609 & 1.2496 & 1.2987 & 1.3814 & 0.1796 & 0.1958 & 0.2094 \\
\rowcolor{gray!10}
\textbf{OURS (MAEDE)}           & \textbf{0.1238} & \textbf{0.1304} & \textbf{0.1543} & \textbf{1.2008} & \textbf{1.2103} & \textbf{1.3204} & \textbf{0.1717} & \textbf{0.1895} & \textbf{0.1956} \\
\bottomrule
\end{tabular}
\label{tab:ablation_length_city}
\vspace{-3mm}
\end{table*}

\begin{table}[tb]
\centering
\vspace{-2mm}
\footnotesize
\caption{Comparison of model size, runtime, and memory consumption under training and inference phases.}
\vspace{-2mm}
\setlength{\tabcolsep}{4.5pt}
\begin{tabular}{l c cc cc}
\toprule
\multirow{2}{*}{\textbf{Models}} 
& \multirow{2}{*}{\textbf{Params}}
& \multicolumn{2}{c}{\textbf{Training}} 
& \multicolumn{2}{c}{\textbf{Inference}} \\
\cmidrule(lr){3-4} \cmidrule(lr){5-6}
& 
& \textbf{Time} 
& \textbf{Mem} 
& \textbf{Time} 
& \textbf{Mem} \\
\midrule
STGCN    & 2.8 M  & 3.81 & 0.95 & 0.76 & 0.41 \\
TimeGrad & 5.1 M  & 6.55 & 1.76 & 1.20 & 1.36 \\
USTD     & 9.8 M  & 11.36 & 3.41 & 1.98 & 2.18 \\
DiffSTG  & 13.8 M   & 32.10 & 4.02 & 2.56 & 2.52 \\
TACTiS   & 15.6 M   & 43.20 & 4.88 & 3.16 & 3.01 \\
CSDI     & 24.5 M   & 26.43 & 3.68 & 2.08 & 2.29 \\
TAD-GNN  & 15.4 M   & 18.90 & 2.74 & 1.35 & 1.62 \\
\textbf{MAEDE} & \textbf{26.3 M} & 29.10 & 4.38 & 2.44 & 2.59 \\
\bottomrule
\end{tabular}
\vspace{-2mm}
\label{tab:complexity_results}
\end{table}

\subsection{Generalization and Robustness}

To evaluate the generalization capability and robustness of MAEDE, we conduct cross-city experiments where models trained on one city’s data are applied to another. Specifically, we perform cross-domain adaptation by training the model on the Changsha dataset and directly deploying it for urban mobility prediction in Guangzhou and Shenzhen and vice versa. The initially trained image encoder, based on satellite imagery from Changsha, is used to extract static environmental features for Guangzhou and Shenzhen. These extracted features are then fused with multi-granularity private car trip data from the target city but processed by the conditional diffusion model trained on other city datasets. {This experiment assesses the adaptability of MAEDE to different urban environments. The results in Fig.~\ref{fig:Generalization} demonstrate that MAEDE maintains high prediction accuracy and reliability, exhibiting strong adaptability to cross-city settings in diverse urban landscapes. Although some performance variations are observed, these differences are attributed to each city’s size and structural characteristics.
}

\subsection{Complexity Analysis}

\begin{figure}[tb]
\centering
\vspace{-2mm}
\subfigure[\textcolor{black}{Consumption of training}]{\includegraphics[width=0.48\linewidth]{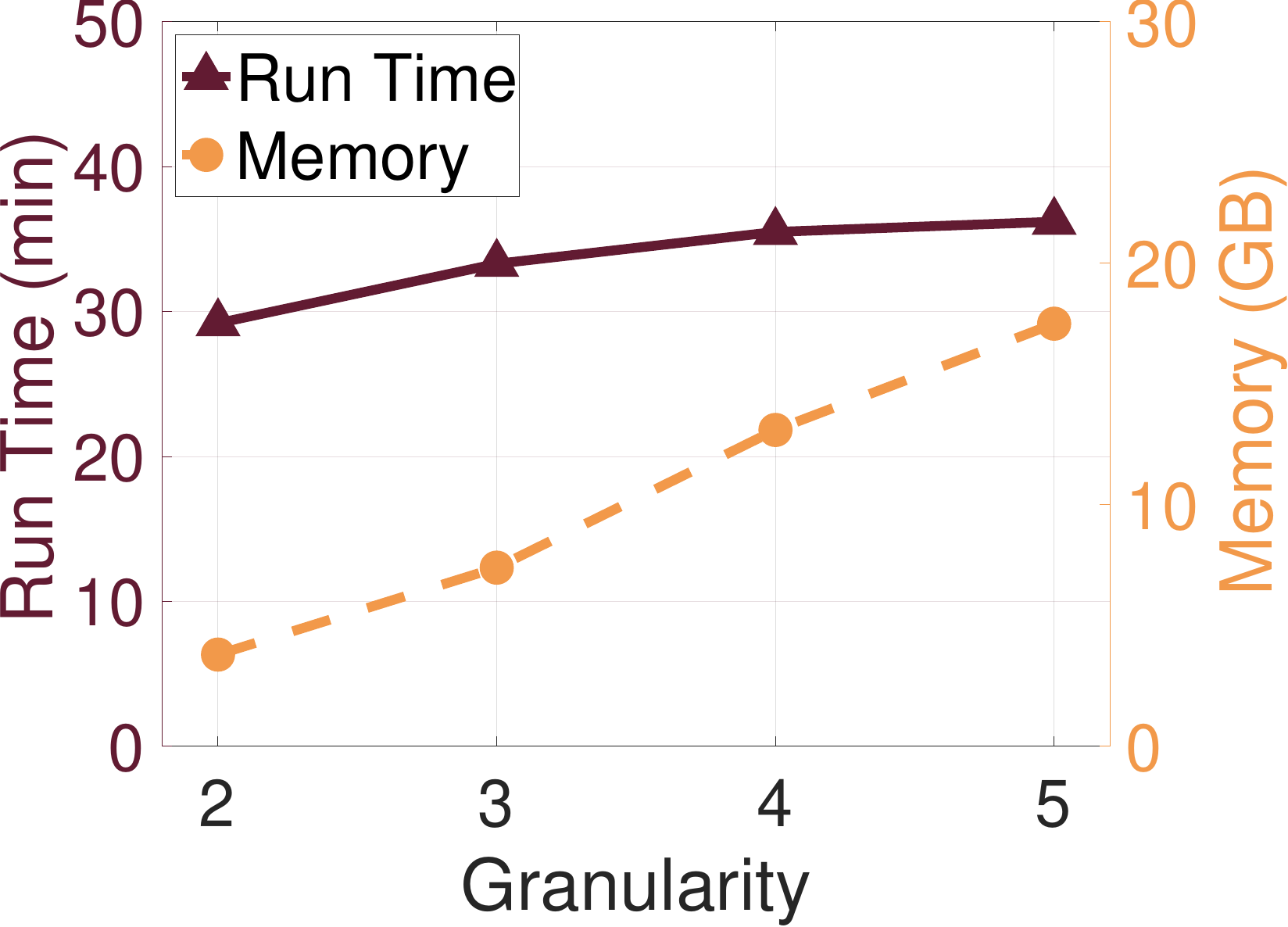}}
\subfigure[\textcolor{black}{Consumption of inferring}]{\includegraphics[width=0.48\linewidth]{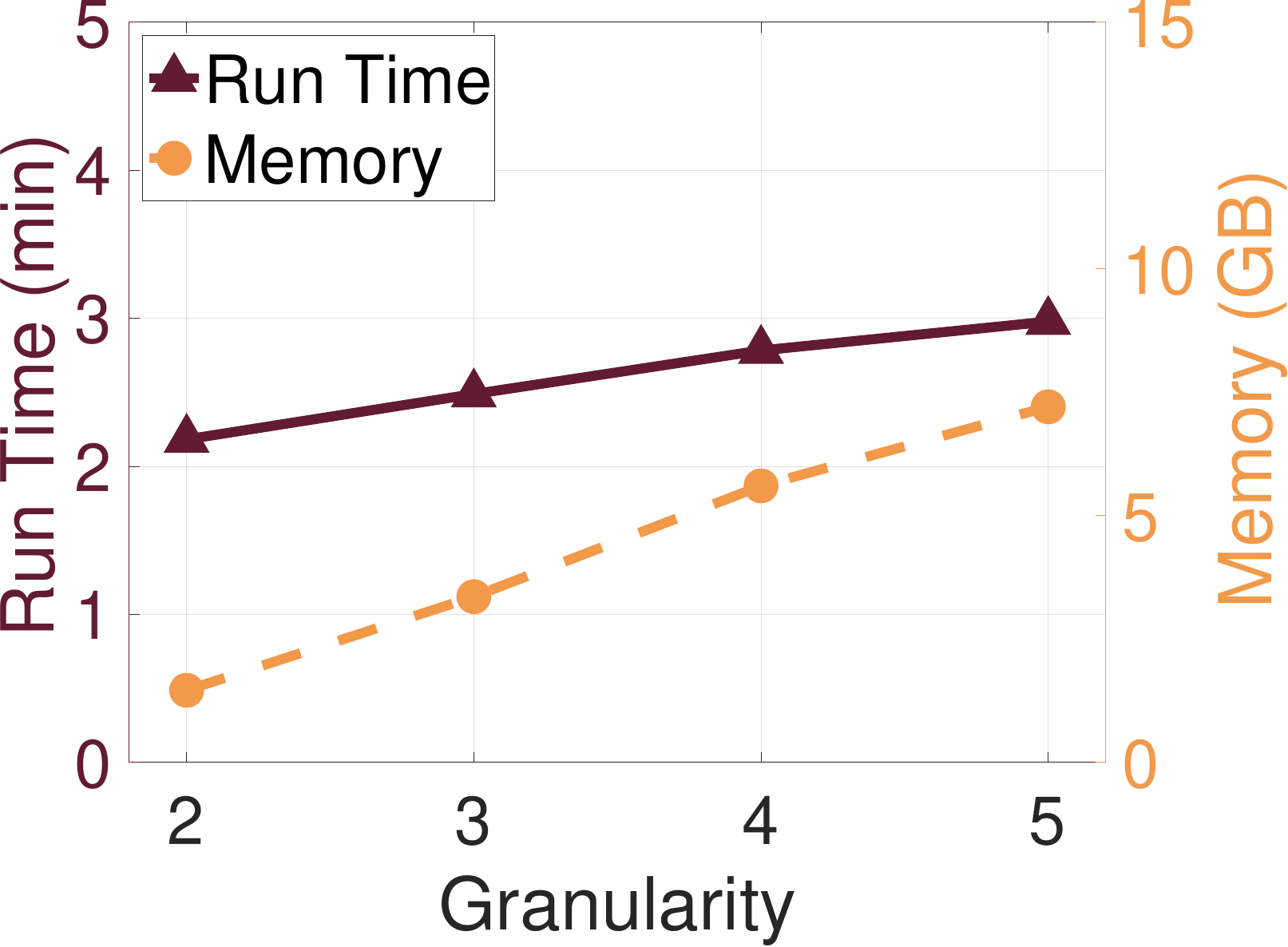}}
\vspace{-2mm}
\caption{Comparison of runtime and memory consumption at different granularity levels.}
\label{fig:complexity}
\vspace{-2mm}
\end{figure}

{
\textcolor{black}{We conducted a detailed empirical evaluation of both training and inference resource consumption to assess the feasibility of MAEDE for real-time applications. Experiments were performed on the Changsha dataset using an NVIDIA GeForce RTX 4090 GPU (24 GB). The approximated model sizes and complexity analysis results are reported in Tab.~\ref{tab:complexity_results}.}
As illustrated in Fig.~\ref{fig:complexity}, both training and inference exhibit a near-linear increase in memory consumption with higher granularity levels, while runtime shows only a moderate rise. These trends are consistent with our model design: each additional granularity introduces an extra RNN unit in the Temporal Process Module and increases computation within the Self-Guiding Diffusion Process Module, resulting in predictable linear growth in complexity. Meanwhile, the Multi-Granularity Data Generator enables parallel forward diffusion across different granularities, effectively mitigating runtime overhead. Consequently, the overall computational complexity remains manageable for real-time or near-real-time applications. Moreover, excessive increases in granularity yield diminishing predictive improvements; therefore, we restrict granularity levels to a practical range, ensuring that memory and computation remain within stable bounds.
}

\section{Related Work}

\subsection{Urban Traffic Prediction}
{
Early methods for traffic forecasting relied on recurrent neural networks such as LSTM~\cite{hochreiter1997lstm}, which effectively modeled temporal dynamics but lacked mechanisms to capture the complex spatial dependencies. To address this limitation, STGCN~\cite{yu2018stgcn} introduced a unified spatio-temporal graph convolutional architecture that jointly captures node-level spatial correlations and temporal patterns. Subsequent works have sought to enhance spatial–temporal representations through more adaptive mechanisms.} Yao et al.~\cite{yao2019} ncorporated spatial–temporal similarity to improve local feature learning, while Guo et al.~\cite{Guo2019} proposed attention-based graph convolutions to better capture dynamic traffic interactions. Beyond temporal modeling, recent studies focus on integrating diverse spatial and contextual cues. Ji et al.~\cite{Ji2024} utilized downsampling attention to extract multi-scale temporal dependencies, Miao et al.~\cite{Miao2024} developed a multi-view channel-wise spatial encoder to handle heterogeneous spatial features, and Patil et al.~\cite{Patil2024} integrated travel-time-aware graph construction to better reflect real-world road network dynamics. Existing methods are deterministic and struggle to model uncertainty or incorporate rich multimodal environmental information.

\subsection{Conditional Diffusion model}
Diffusion models have recently become prominent tools for probabilistic forecasting. Rasul et al.~\cite{rasulAutoregressiveDenoisingDiffusion} introduced TimeGrad, a denoising diffusion model for forecasting multivariate time series. Tashiro et al. proposed CSDI~\cite{tashiro2021csdi} for conditional imputation, and Ho and Salimans~\cite{ho2022classifier}  developed classifier-free guidance to improve generation quality without relying on external classifiers. Dhariwal and Nichol~\cite{nichol2021improved} introduced classifier-based conditional generation, which offers stronger target conditioning at the cost of reduced flexibility. To improve scalability, Tashiro et al.~\cite{tashiro2023tactis} extended transformer-based architectures through TACTiS, thereby enhancing the long-range forecasting capability. Yu et al.~\cite{yu2023freedom} proposed training-free energy-based objectives to reduce optimization complexity, and Zhao et al.~\cite{zhaoadding} incorporated reinforcement learning to enable finer-grained conditional control. Tang et al.~\cite{tang2024conditional} and Gu et al.~\cite{gu2024kaleido} further refined conditional diffusion via manifold-adaptive estimation and autoregressive latent modeling, respectively. {In the spatio-temporal domain, Wen et al. presented DiffSTG [29], integrating diffusion with graph learning, and Hu et al.~\cite{hu2024towards} proposed a unified framework for probabilistic spatio-temporal modeling. While these models demonstrate strong generative capabilities and support for uncertainty modeling, they typically assume fixed temporal structures or lack mechanisms for hierarchical guidance and multimodal context integration.}

\section{Conclusion}
\label{sec:con}
In this paper, we modeled private car activity using ASL sequences and investigated the problem of predicting regional private car activity. To resolve this problem, we propose MAEDE (Multi-grAnularity sElf-guiding Diffusion modEl), a novel approach for probabilistic forecasting of regional private car activity. By leveraging a conditional diffusion model, MAEDE effectively captures complex joint distributions. Furthermore, integrating satellite imagery with an LLM-enhanced CLIP enables MAEDE to incorporate environmental features more effectively. To tackle both long-term trends and short-term fluctuations, MAEDE employs a multi-granularity self-guiding mechanism to guide the denoising process, ensuring the simultaneous extraction of high-level patterns and fine-grained details. Extensive experiments on real-world private car trip datasets demonstrate that MAEDE significantly outperforms state-of-the-art baselines and exhibits robustness and generalization across diverse urban environments. Our research sheds new light on extracting hidden spatiotemporal dependencies and aids in transportation planning and smart city development.

\textcolor{black}{
The proposed probabilistic forecasting framework aligns with the principles of Calm Mobility~\cite{turon2025sustainable}, emphasizing predictability, comfort, and reduced travel stress in urban transportation systems. By enabling smoother traffic coordination and adaptive urban planning, MAEDE contributes to the development of user-centered, seamless, and resilient mobility environments. Beyond its technical contributions, this work also connects with broader discussions in urban mobility research and planning theory, where predictive analytics are increasingly used to support sustainable and human-centered city design~\cite{Zhou2022, sheller2006new, banister2008sustainable}. The ability to anticipate regional mobility dynamics provides planners with quantitative evidence for interventions in parking allocation, congestion mitigation, and multimodal transport coordination—key elements in realizing the vision of “calm,” efficient, and equitable mobility~\cite{galanakis2024place, sonnberger2021sociocultural, xiao2020extracting}. In this sense, MAEDE not only advances spatiotemporal forecasting techniques but also reinforces the interdisciplinary integration between intelligent transportation systems, urban planning, and mobility behavior studies, bridging technical modeling with policy-relevant insights for future smart cities.
}

\bibliographystyle{IEEEtran}
\bibliography{bibliography}

@article{yiche2024,
  author = {Yiche News},
  title = {Comparison of car ownership between China and the United States: The number of cars in the United States is 280 million, how many are in China?},
  year = {2024},
  url = {https://news.yiche.com/hao/wenzhang/87587805/}
}

@article{NBS2021,
  author = {{National Bureau of Statistics of China}},
  title = {China Statistical Yearbook 2021},
  year = {2021},
  howpublished = {Online},
  url = {http://www.stats.gov.cn/tjsj/ndsj/2021/indexch.htm}
}

@article{xiao2025llm,
  title={An LLM-Enhanced Conditional Diffusion Model for Mobile Traffic Prediction},
  author={Xiao, Zhu and Wang, Rui and Bai, Jing and Li, Tong and Zhang, Shiyuan and Li, Keqin and Han, Zhu},
  journal={IEEE Communications Magazine},
  volume={63},
  number={9},
  pages={60--67},
  year={2025},
  publisher={IEEE}
}

@article{zhang2022cityneuro,
  title={Cityneuro: Towards location and time prediction for urban abnormal events},
  author={Zhang, Mingyang and Li, Tong and Hui, Pan},
  journal={IEEE Transactions on Knowledge and Data Engineering},
  volume={35},
  number={12},
  pages={12125--12138},
  year={2022},
  publisher={IEEE}
}

@inproceedings{li2019driving,
  title={Driving big data: A first look at driving behavior via a large-scale private car dataset},
  author={Li, Tong and Alhilal, Ahmad and Zhang, Anlan and Hoque, Mohammad A and Chatzopoulos, Dimitris and Xiao, Zhu and Li, Yong and Hui, Pan},
  booktitle={2019 IEEE 35th international conference on data engineering workshops (ICDEW)},
  pages={61--68},
  year={2019},
  organization={IEEE}
}

@article{liu2025vehicle,
  title={Vehicle trajectory data processing, analytics, and applications: A survey},
  author={Liu, Chenxi and Xiao, Zhu and Long, Wangchen and Li, Tong and Jiang, Hongbo and Li, Keqin},
  journal={ACM Computing Surveys},
  volume={57},
  number={9},
  pages={1--36},
  year={2025},
  publisher={ACM New York, NY}
}

@inproceedings{zhang2021multi,
  title={Multi-view joint graph representation learning for urban region embedding},
  author={Zhang, Mingyang and Li, Tong and Li, Yong and Hui, Pan},
  booktitle={Proceedings of the twenty-ninth international conference on international joint conferences on artificial intelligence},
  pages={4431--4437},
  year={2021}
}

@article{fang2023learning,
  author={Fang, Hui and Xiao, Zhu and Zheng, Pengfei and Chen, Hongyang and Li, Zhao and Bu, Jiajun and Wang, Haishuai},
  journal={IEEE Transactions on Mobile Computing}, 
  title={Learning Co-occurrence Patterns for Next Destination Recommendation}, 
  year={2024},
  volume={23},
  number={6},
  pages={7225-7237},
  doi={10.1109/TMC.2023.3333944}}

@article{speizer2024integrated,
  title={Integrated assessment modeling of a zero-emissions global transportation sector},
  author={Speizer, Simone and Fuhrman, Jay and Aldrete Lopez, Laura and George, Mel and Kyle, Page and Monteith, Seth and McJeon, Haewon},
  journal={Nature Communications},
  volume={15},
  number={1},
  pages={4439},
  year={2024},
  publisher={Nature Publishing Group UK London}
}

@article{xiao2023parking,
  title={Parking prediction in smart cities: A survey},
  author={Xiao, Xiao and Peng, Ziyan and Lin, Yunqing and Jin, Zhiling and Shao, Wei and Chen, Rui and Cheng, Nan and Mao, Guoqiang},
  journal={IEEE Transactions on Intelligent Transportation Systems},
  volume={24},
  number={10},
  pages={10302--10326},
  year={2023},
  publisher={IEEE}
}

@article{hu2023spatio,
  title={Spatio-temporal trajectory similarity measures: A comprehensive survey and quantitative study},
  author={Hu, Danlei and Chen, Lu and Fang, Hanxi and Fang, Ziquan and Li, Tianyi and Gao, Yunjun},
  journal={IEEE Transactions on Knowledge and Data Engineering},
  volume={36},
  number={5},
  pages={2191--2212},
  year={2023},
  publisher={IEEE}
}

@article{diao2021impacts,
  title={Impacts of transportation network companies on urban mobility},
  author={Diao, Mi and Kong, Hui and Zhao, Jinhua},
  journal={Nature Sustainability},
  volume={4},
  number={6},
  pages={494--500},
  year={2021},
  publisher={Nature Publishing Group UK London}
}

@article{zhao2022urban,
  author = {Jianming Zhao and Yingjia Tan and Yizhou Zhu},
  title = {Research on Planning of Urban Rail Connection with Conventional Bus},
  journal = {Open Journal of Transportation Technologies},
  volume = {11},
  number = {3},
  pages = {193--202},
  year = {2022}
}

@article{yan2024bus,
  author = {Minzu Yan and Guanpeng Dong and Binbin Lu},
  title = {Bus-Subway Interchange Mode Research with IC Card Data},
  journal = {Journal of Geo-Information Science},
  volume = {26},
  number = {6},
  pages = {1351--1362},
  year = {2024}
}

@article{feng2017role,
  author = {Yesu Feng},
  title = {The Role of Sharing Traffic in Urban Traffic Development},
  journal = {Urban Development Studies},
  volume = {24},
  number = {6},
  pages = {C9--C12},
  year = {2017},
}

@article{hochreiter1997lstm,
  author    = {Sepp Hochreiter and J{\"u}rgen Schmidhuber},
  title     = {Long Short-Term Memory},
  journal   = {Neural Computation},
  year      = {1997},
  volume    = {9},
  number    = {8},
  pages     = {1735--1780},
  doi       = {10.1162/neco.1997.9.8.1735},
  publisher = {MIT Press}
}

@inproceedings{huUnifyingDiffusionModels2024,
  title = {Towards Unifying Diffusion Models for Probabilistic Spatio-Temporal Graph Learning},
  booktitle = {Proceedings of the 32nd ACM International Conference on Advances in Geographic Information Systems},
  author = {Hu, Junfeng and Liu, Xu and Fan, Zhencheng and Liang, Yuxuan and Zimmermann, Roger},
  year = {2024},
  month = oct,
  pages = {135--146},
  publisher = {ACM},
  address = {Atlanta GA USA},
  urldate = {2025-01-14},
  isbn = {9798400711077},
  langid = {english}
}

@article{li2023learning,
  title={Learning representations of satellite imagery by leveraging point-of-interests},
  author={Li, Tong and Xi, Yanxin and Wang, Huandong and Li, Yong and Tarkoma, Sasu and Hui, Pan},
  journal={ACM Transactions on Intelligent Systems and Technology},
  volume={14},
  number={4},
  pages={1--32},
  year={2023},
  publisher={ACM New York, NY}
}

@inproceedings{xi2022beyond,
  title={Beyond the first law of geography: Learning representations of satellite imagery by leveraging point-of-interests},
  author={Xi, Yanxin and Li, Tong and Wang, Huandong and Li, Yong and Tarkoma, Sasu and Hui, Pan},
  booktitle={Proceedings of the ACM web conference 2022},
  pages={3308--3316},
  year={2022}
}

@inproceedings{long2023practical,
  title={Practical synthetic human trajectories generation based on variational point processes},
  author={Long, Qingyue and Wang, Huandong and Li, Tong and Huang, Lisi and Wang, Kun and Wu, Qiong and Li, Guangyu and Liang, Yanping and Yu, Li and Li, Yong},
  booktitle={Proceedings of the 29th ACM SIGKDD conference on knowledge discovery and data mining},
  pages={4561--4571},
  year={2023}
}

@article{wenDiffSTGProbabilisticSpatioTemporal2024,
  title = {DiffSTG: Probabilistic Spatio-Temporal Graph Forecasting with Denoising Diffusion Models},
  author={Wen, Haomin and Lin, Youfang and Xia, Yutong and Wan, Huaiyu and Wen, Qingsong and Zimmermann, Roger and Liang, Yuxuan},
  booktitle={Proceedings of the 31st ACM international conference on advances in geographic information systems},
  pages={1--12},
  year={2023}
}

@article{xiaoPredictingUrbanRegion2023,
  title = {Predicting Urban Region Heat via Learning Arrive-Stay-Leave Behaviors of Private Cars},
  author = {Xiao, Zhu and Li, Hao and Jiang, Hongbo and Li, You and Alazab, Mamoun and Zhu, Yongdong and Dustdar, Schahram},
  year = {2023},
  month = oct,
  journal = {IEEE Transactions on Intelligent Transportation Systems},
  volume = {24},
  number = {10},
  pages = {10843--10856}
}

@inproceedings{yu2018stgcn,
  author    = {Bing Yu and Haoteng Yin and Zhanxing Zhu},
  title     = {Spatio-Temporal Graph Convolutional Networks: A Deep Learning Framework for Traffic Forecasting},
  booktitle = {Proceedings of the 27th International Joint Conference on Artificial Intelligence (IJCAI)},
  year      = {2018},
  pages     = {3634--3640},
  doi       = {10.24963/ijcai.2018/505}
}

@inproceedings{Yao2019,
  author    = {Huaxiu Yao and Ying Wei and Xianfeng Tang and Zheng G. Li and Zhenhui Li},
  title     = {Revisiting Spatial-Temporal Similarity: A Deep Learning Framework for Traffic Prediction},
  booktitle = {Proceedings of the AAAI Conference on Artificial Intelligence},
  volume    = {33},
  number    = {01},
  pages     = {5668--5675},
  year      = {2019}
}

@article{Guo2019,
  author    = {Shengnan Guo and Youfang Lin and Ning Feng and Chao Song and Huaiyu Wan},
  title     = {Attention Based Spatial-Temporal Graph Convolutional Networks for Traffic Flow Forecasting},
  journal   = {Proceedings of the AAAI Conference on Artificial Intelligence},
  volume    = {33},
  number    = {01},
  pages     = {922--929},
  year      = {2019}
}

@article{Zhou2022,
  author    = {Siji Zhou and Zhenkun Qian},
  title     = {High-Precision Traffic Flow Big Data Prediction Based on Deep Learning},
  journal   = {Journal of Southwest China Normal University (Natural Science Edition)},
  volume    = {47},
  number    = {8},
  pages     = {1--8},
  year      = {2022}
}

@article{Ji2024,
  author    = {Xilai Ji and Yutong Huang and Shuang Leng and Aihua Zhang},
  title     = {Traffic Flow Prediction Based on Downsampled Interactive Spatio-Temporal Attention Network},
  journal   = {Modeling and Simulation},
  volume    = {13},
  number    = {4},
  pages     = {4090--4103},
  year      = {2024}
}

@article{Miao2024,
  title={Deep Multi-View Channel-Wise Spatio-Temporal Network for Traffic Flow Prediction},
  author={Miao, Hao and Wang, Senzhang and Zhang, Meiyue and Guo, Diansheng and Sun, Funing and Yang, Fan},
  journal={CoRR},
  year={2024}
}

@inproceedings{Patil2024,
  title={Urban Traffic Forecasting with Integrated Travel Time and Data Availability in a Conformal Graph Neural Network Framework},
  author={Patil, Mayur and Ahmed, Qadeer and Midlam-Mohler, Shawn},
  booktitle={2024 IEEE 27th International Conference on Intelligent Transportation Systems (ITSC)},
  pages={2482--2487},
  year={2024},
  organization={IEEE}
}

@inproceedings{ho2022classifier,
  title={Classifier-Free Diffusion Guidance},
  author={Ho, Jonathan and Salimans, Tim},
  booktitle={NeurIPS 2021 Workshop on Deep Generative Models and Downstream Applications},
  year={2021}
}

@inproceedings{zhaoadding,
  title={Adding Conditional Control to Diffusion Models with Reinforcement Learning},
  author={Zhao, Yulai and Uehara, Masatoshi and Scalia, Gabriele and Kung, Sunyuan and Biancalani, Tommaso and Levine, Sergey and Hajiramezanali, Ehsan},
  booktitle={The Thirteenth International Conference on Learning Representations},
  year={2026}
}

@inproceedings{yu2023freedom,
  title={Freedom: Training-free energy-guided conditional diffusion model},
  author={Yu, Jiwen and Wang, Yinhuai and Zhao, Chen and Ghanem, Bernard and Zhang, Jian},
  booktitle={Proceedings of the IEEE/CVF International Conference on Computer Vision},
  pages={23174--23184},
  year={2023}
}

@inproceedings{tang2024conditional,
  title={Conditional Diffusion Models are Minimax-Optimal and Manifold-Adaptive for Conditional Distribution Estimation},
  author={Tang, Rong and Lin, Lizhen and Yang, Yun},
  booktitle={The Thirteenth International Conference on Learning Representations},
  year={2025}
}

@article{gu2024kaleido,
  title={Kaleido diffusion: Improving conditional diffusion models with autoregressive latent modeling},
  author={Gu, Jiatao and Shen, Ying and Zhai, Shuangfei and Zhang, Yizhe and Jaitly, Navdeep and Susskind, Joshua},
  journal={Advances in Neural Information Processing Systems},
  volume={37},
  pages={5498--5527},
  year={2024}
}

@article{ho2020denoising,
  title={Denoising diffusion probabilistic models},
  author={Ho, Jonathan and Jain, Ajay and Abbeel, Pieter},
  journal={Advances in Neural Information Processing Systems},
  volume={33},
  pages={6840--6851},
  year={2020}
}

@article{nichol2021improved,
  title={Improved denoising diffusion probabilistic models},
  author={Nichol, Alex and Dhariwal, Prafulla},
  journal={International Conference on Machine Learning (ICML)},
  year={2021}
}

@article{zhang2022toward,
  title={Toward Predicting Stay Time for Private Car Users: A {RNN-NALU} Approach},
  author={Zhang, Qibo and Zeng, Fanzi and Xiao, Zhu and Jiang, Hongbo and Regan, Amelia C. and Yang, Kehua and Zhu, Yongdong},
  journal={IEEE Transactions on Vehicular Technology},
  volume={71},
  number={6},
  pages={6007--6018},
  year={2022},
  publisher={IEEE}
}

@article{jin2022survey,
  title={A survey and experimental study on privacy-preserving trajectory data publishing},
  author={Jin, Fengmei and Hua, Wen and Francia, Matteo and Chao, Pingfu and Orlowska, Maria E and Zhou, Xiaofang},
  journal={IEEE Transactions on Knowledge and Data Engineering},
  volume={35},
  number={6},
  pages={5577--5596},
  year={2022},
  publisher={IEEE}
}

@article{li2014stochastic,
  title = {Research Progress on Stochastic Behavior of Traffic Flow},
  author = {Li, Wei and Zhang, Wei and Wang, Wei},
  journal = {Journal of Transportation Systems Engineering and Information Technology},
  volume = {14},
  number = {5},
  pages = {1--8},
  year = {2014},
  publisher = {Elsevier}
}

@inproceedings{zhao2016urban,
  title={Urban human mobility data mining: An overview},
  author={Zhao, Kai and Tarkoma, Sasu and Liu, Siyuan and Vo, Huy},
  booktitle={2016 IEEE International Conference on Big Data (Big Data)},
  pages={1911--1920},
  year={2016},
  organization={IEEE}
}

@article{yuan2014discovering,
  title={Discovering urban functional zones using latent activity trajectories},
  author={Yuan, Nicholas Jing and Zheng, Yu and Xie, Xing and Wang, Yingzi and Zheng, Kai and Xiong, Hui},
  journal={IEEE Transactions on Knowledge and Data Engineering},
  volume={27},
  number={3},
  pages={712--725},
  year={2014},
  publisher={IEEE}
}

@inproceedings{zhang2023deep,
  title={Deep transfer learning for city-scale cellular traffic generation through urban knowledge graph},
  author={Zhang, Shiyuan and Li, Tong and Hui, Shuodi and Li, Guangyu and Liang, Yanping and Yu, Li and Jin, Depeng and Li, Yong},
  booktitle={Proceedings of the 29th ACM SIGKDD conference on knowledge discovery and data mining},
  pages={4842--4851},
  year={2023}
}

@article{chen2018tripimputor,
  title={TripImputor: Real-time imputing taxi trip purpose leveraging multi-sourced urban data},
  author={Chen, Chao and Jiao, Shuhai and Zhang, Shu and Liu, Weichen and Feng, Liang and Wang, Yasha},
  journal={IEEE Transactions on Intelligent Transportation Systems},
  volume={19},
  number={10},
  pages={3292--3304},
  year={2018},
  publisher={IEEE}
}

@article{liu2021understanding,
  title={Understanding the Regular Travel Behavior of Private Vehicles: An Empirical Evaluation and A Semi-supervised Model},
  author={Liu, Chenxi and Cai, Jiawei and Wang, Dong and Tang, Jiaxin and Wang, Lei and Chen, Huiling and Xiao, Zhu},
  journal={IEEE Sensors Journal},
  volume={21},
  number={17},
  pages={19078--19090},
  year={2021},
  publisher={IEEE}
}

@article{wan2018smopat,
  title={SMOPAT: Mining semantic mobility patterns from trajectories of private vehicles},
  author={Wan, Chengcheng and Zhu, Yanmin and Yu, Jiadi and Shen, Yanyan},
  journal={Information Sciences},
  volume={429},
  pages={12--25},
  year={2018},
  publisher={Elsevier}
}

@article{cao2007discovery,
  title={Discovery of periodic patterns in spatiotemporal sequences},
  author={Cao, Huiping and Mamoulis, Nikos and Cheung, David W},
  journal={IEEE Transactions on Knowledge and Data Engineering},
  volume={19},
  number={4},
  pages={453--467},
  year={2007},
  publisher={IEEE}
}

@inproceedings{rasulAutoregressiveDenoisingDiffusion,
  title={Autoregressive denoising diffusion models for multivariate probabilistic time series forecasting},
  author={Rasul, Kashif and Seward, Calvin and Schuster, Ingmar and Vollgraf, Roland},
  booktitle={International Conference on Machine Learning},
  pages={8857--8868},
  year={2021},
  organization={PMLR}
}

@article{wang2019revealing,
  title = {Revealing Spatial-Temporal Characteristics and Patterns of Urban Travel: A Large-Scale Analysis and Visualization Study with Taxi GPS Data},
  author = {Wang, Huihui and Huang, Hong and Ni, Xiaoyong and Zeng, Weihua},
  journal = {ISPRS International Journal of Geo-Information},
  volume = {8},
  number = {6},
  pages = {257},
  year = {2019},
  publisher = {MDPI},
}

@inproceedings{song2021scorebased,
  title={Conditional Image Generation with Score-Based Diffusion Models},
  author={Song, Yang and Sohl-Dickstein, Jascha and Kingma, Diederik P and Kumar, Abhishek and Ermon, Stefano and Poole, Ben},
  booktitle={Advances in Neural Information Processing Systems},
  volume={34},
  pages={17284--17294},
  year={2021}
}

@inproceedings{liu2023pristi,
  title={PriSTI: A Conditional Diffusion Framework for Spatiotemporal Imputation},
  author={Liu, Mingzhe and Huang, Han and Feng, Hao and Sun, Leilei and Du, Bowen and Fu, Yanjie},
  booktitle={Proceedings of the 30th ACM International Conference on Information and Knowledge Management (CIKM)},
  pages={1234--1243},
  year={2023}
}

@inproceedings{hu2024towards,
  title={Towards unifying diffusion models for probabilistic spatio-temporal graph learning},
  author={Hu, Junfeng and Liu, Xu and Fan, Zhencheng and Liang, Yuxuan and Zimmermann, Roger},
  booktitle={Proceedings of the 32nd ACM International Conference on Advances in Geographic Information Systems},
  pages={135--146},
  year={2024}
}

@inproceedings{ester1996density,
  title={A Density-Based Algorithm for Discovering Clusters in Large Spatial Databases with Noise},
  author={Ester, Martin and Kriegel, Hans-Peter and Sander, J{\"o}rg and Xu, Xiaowei},
  booktitle={Proceedings of the Second International Conference on Knowledge Discovery and Data Mining (KDD-96)},
  pages={226--231},
  year={1996},
  organization={AAAI Press}
}

@inproceedings{li2022predicting,
  title={Predicting Multi-level Socioeconomic Indicators from Structural Urban Imagery},
  author={Li, Tong and Xin, Shiduo and Xi, Yanxin and Tarkoma, Sasu and Li, Yong},
  booktitle={Proceedings of the 30th ACM International Conference on Advances in Geographic Information Systems},
  pages={1--10},
  year={2022},
  organization={ACM}
}

@inproceedings{tashiro2023tactis,
  title={TACTiS: Transformers for Probabilistic Time Series Forecasting},
  author={Tashiro, Yusuke and Zhang, Xuechen and Wang, Ziyan and Song, Yang},
  booktitle={International Conference on Learning Representations (ICLR)},
  year={2023}
}

@article{tashiro2021csdi,
  title={{CSDI}: Conditional score-based diffusion models for probabilistic time series imputation},
  author={Tashiro, Yusuke and Song, Jiaming and Song, Yang and Ermon, Stefano},
  journal={Advances in neural information processing systems},
  volume={34},
  pages={24804--24816},
  year={2021}
}

@article{mesfin2024urban,
  title={Urban traffic-parking system dynamics model with macroscopic properties: a comparative study between Shanghai and Zurich},
  author={Mesfin, Biruk Gebremedhin and Li, Zihao and Sun, Daniel Jian and Chen, Deming and Xi, Yueting},
  journal={Humanities and Social Sciences Communications},
  volume={11},
  number={1},
  pages={1--13},
  year={2024},
  publisher={Palgrave}
}

@article{xiao2024exploring,
  title={Exploring spatio-temporal carbon emission across passenger car trajectory data},
  author={Xiao, Zhu and Liu, Bo and Wu, Linshan and Jiang, Hongbo and Xia, Beihao and Li, Tao and Wang, Cassandra C},
  journal={IEEE Transactions on Intelligent Transportation Systems},
  year={2024},
  publisher={IEEE}
}

@article{xiao2021trajectory,
  title={Trajectory data acquisition via private car positioning based on tightly-coupled GPS/OBD integration in urban environments},
  author={Xiao, Zhu and Chen, Yanxun and Alazab, Mamoun and Chen, Hongyang},
  journal={IEEE Transactions on Intelligent Transportation Systems},
  volume={23},
  number={7},
  pages={9680--9691},
  year={2021},
  publisher={IEEE}
}

@article{turon2025sustainable,
  title={Sustainable Urban Mobility Transitions—From Policy Uncertainty to the CalmMobility Paradigm},
  author={Turo{\'n}, Katarzyna},
  journal={Smart Cities},
  volume={8},
  number={5},
  pages={164},
  year={2025},
  publisher={MDPI}
}

@article{sheller2006new,
  title={The new mobilities paradigm},
  author={Sheller, Mimi and Urry, John},
  journal={Environment and planning A},
  volume={38},
  number={2},
  pages={207--226},
  year={2006},
  publisher={Sage Publications Sage UK: London, England}
}

@article{galanakis2024place,
  title={Place-based sustainable urban mobility: a conceptual framework to spark local designs},
  author={Galanakis, Kostas and Heinz, Helen and Marggraf, Clemens},
  journal={Regional Studies},
  volume={58},
  number={12},
  pages={2419--2434},
  year={2024},
  publisher={Taylor \& Francis}
}

@article{sonnberger2021sociocultural,
  title={Sociocultural dimensions of mobility transitions to come: introduction to the special issue},
  author={Sonnberger, Marco and Graf, Antonia},
  journal={Sustainability: Science, Practice and Policy},
  volume={17},
  number={1},
  pages={173--184},
  year={2021},
  publisher={Taylor \& Francis}
}

@article{xiao2020extracting,
  title={On extracting regular travel behavior of private cars based on trajectory data analysis},
  author={Xiao, Zhu and Xu, Shenyuan and Li, Tao and Jiang, Hongbo and Zhang, Rui and Regan, Amelia C and Chen, Hongyang},
  journal={IEEE Transactions on Vehicular Technology},
  volume={69},
  number={12},
  pages={14537--14549},
  year={2020},
  publisher={IEEE}
}

@article{banister2008sustainable,
  title={The sustainable mobility paradigm},
  author={Banister, David},
  journal={Transport policy},
  volume={15},
  number={2},
  pages={73--80},
  year={2008},
  publisher={Elsevier}
}

\vskip -2\baselineskip plus -1fil

\begin{IEEEbiography}
[{\includegraphics[width=1in,height=1.25in,clip,keepaspectratio]{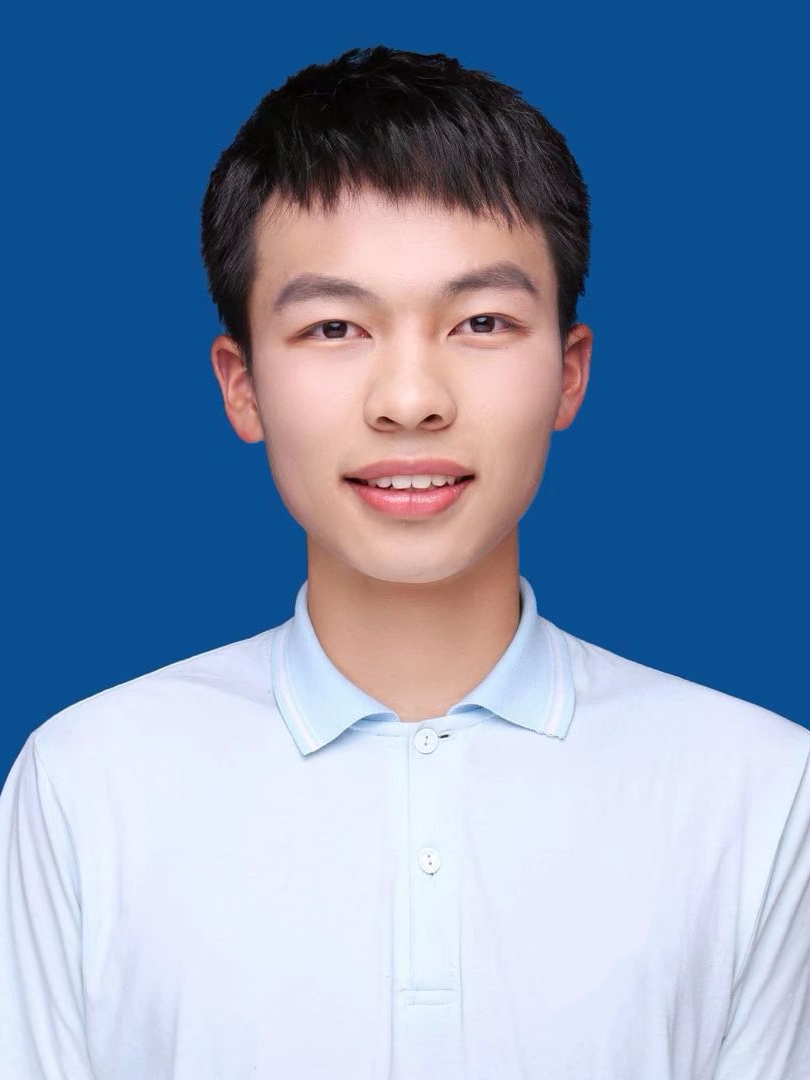}}] {Bo Liu} is currently pursuing the M.S. degree in software engineering with Hunan University, Changsha, China. His research interests include intelligent transportation systems, satellite navigation, and big data mining.
\end{IEEEbiography}

\vskip -2\baselineskip plus -1fil

\begin{IEEEbiography}
[{\includegraphics[width=1in,height=1.25in,clip,keepaspectratio]{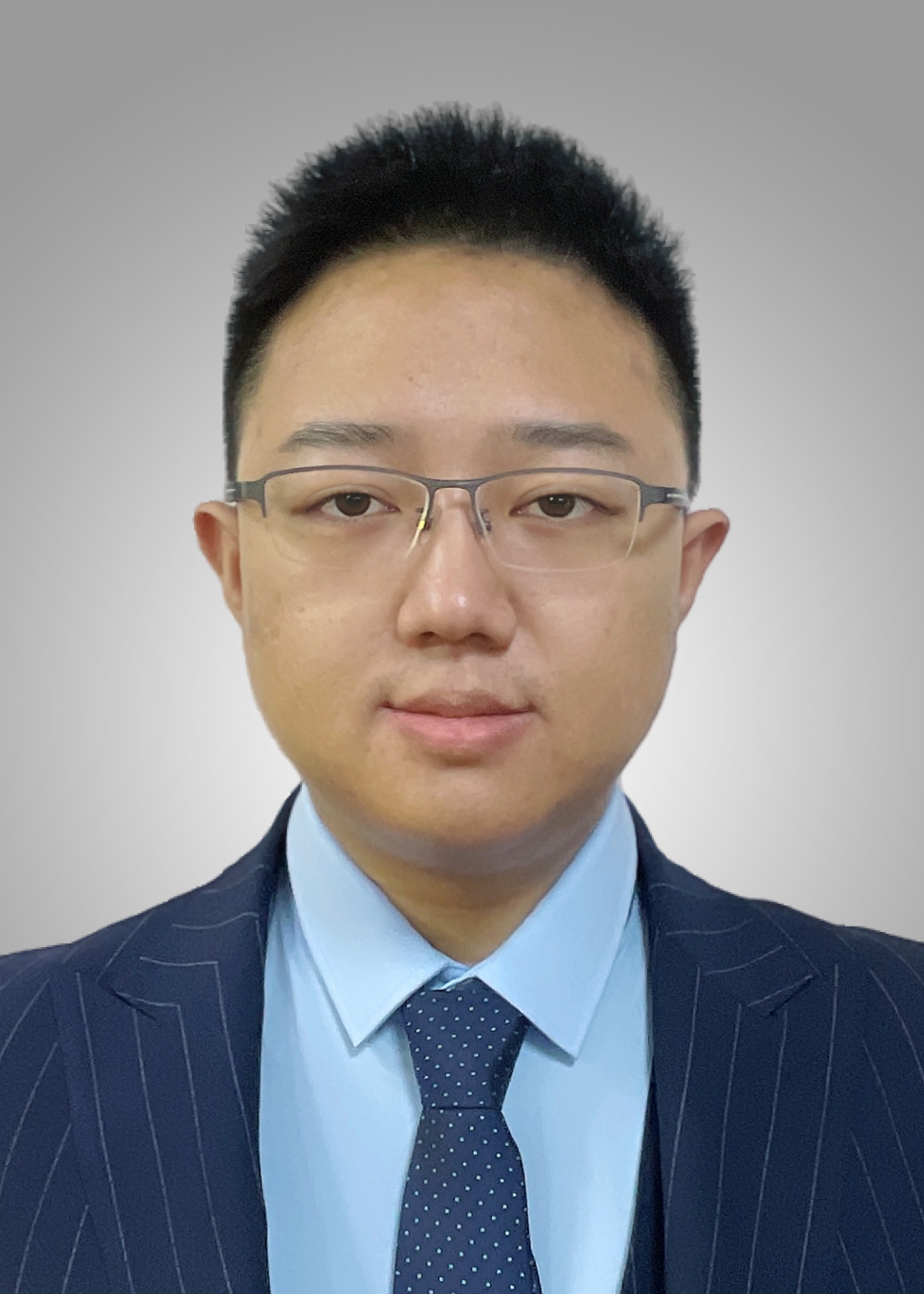}}] {Tong Li} (Member, IEEE) received the Ph.D. in Computer Science and Engineering from the Hong Kong University of Science and Technology in 2021 and the Ph.D. in Computer Science from the University of Helsinki in 2022. He is currently a Full Professor at the College of Computer Science and Electronic Engineering, Hunan University, Changsha, China. His research interests include mobile computing, network digital twin, and spatial intelligence. 
\end{IEEEbiography}

\vskip -2\baselineskip plus -1fil

\begin{IEEEbiography}[{\includegraphics[width=1in,height=1.25in,clip,keepaspectratio]{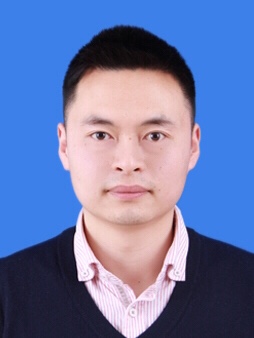}}]{Zhu Xiao} (M'12-SM'19) received the M.S. and Ph.D. degrees in communication and information systems from Xidian University, China, in 2007 and 2009, respectively. From 2010 to 2012, he was a Research Fellow with the Department of Computer Science and Technology, University of Bedfordshire, U.K. He is currently a Full Professor with the College of Computer Science and Electronic Engineering, Hunan University, China. His research interests include intelligent information processing, Internet of Vehicles and wireless communications systems. He is a senior member of the IEEE and serving as an Associated Editor for IEEE Transactions on Intelligent Transportation Systems.
\end{IEEEbiography}

\vskip -2\baselineskip plus -1fil

\begin{IEEEbiography}[{\includegraphics[width=1in,height=1.25in,clip,keepaspectratio]{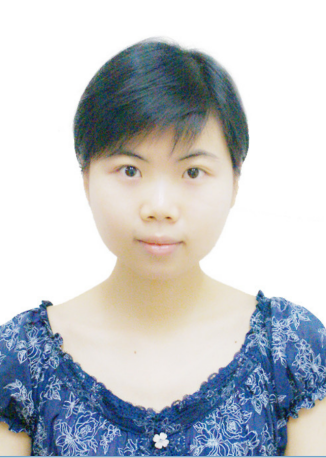}}]{Wenjie Chen} is a professor at the School of Economics and Management, Central South University of Forestry and Technology, China. She received her M.S. degree in Management from Hunan University, China, in 2009, and her Ph.D. in Business Administration from the same university in 2013. Her research interests include low-carbon transition, urban resilience, and technological innovation.
\end{IEEEbiography}

\vskip -2\baselineskip plus -1fil

\begin{IEEEbiography}[{\includegraphics[width=1in,height=1.25in,clip,keepaspectratio]{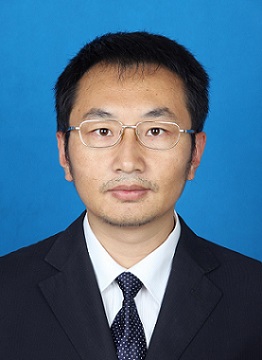}}]{Yongqiang Hei} received the B.E. and Ph.D. degrees in telecommunication engineering from Xidian University, Xi’an, China, in 2006 and 2010, respectively.He is currently a Professor with the State Key Laboratory of Integrated Services Networks, Xidian University. His research interests include 5G/6G communication, massive MIMO, green communication, beamforming and array antennas.
\end{IEEEbiography}

\vskip -2\baselineskip plus -1fil

\begin{IEEEbiography}[{\includegraphics[width=1in,height=1.25in,clip,keepaspectratio]{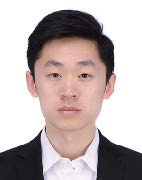}}]{Jingtao Ding} received the BS degrees in electronic engineering and the PhD degree in electronic engineering from Tsinghua University, Beijing, China, in 2015 and 2020, respectively. He is currently a postdoctoral research fellow with the Department of Electronic Engineering, Tsinghua University. His research interests include mobile computing, spatiotemporal data mining and user behavior modeling. He has more than 60 publications in journals and conferences such as IEEE Transactions on Knowledge and Data Engineering, ACM Transactions on Information Systems, KDD, NeurIPS, WWW, ICLR, SIGIR, IJCAI, etc.
\end{IEEEbiography}

\vskip -2\baselineskip plus -1fil

\begin{IEEEbiography}[{\includegraphics[width=1in,height=1.25in,clip,keepaspectratio]{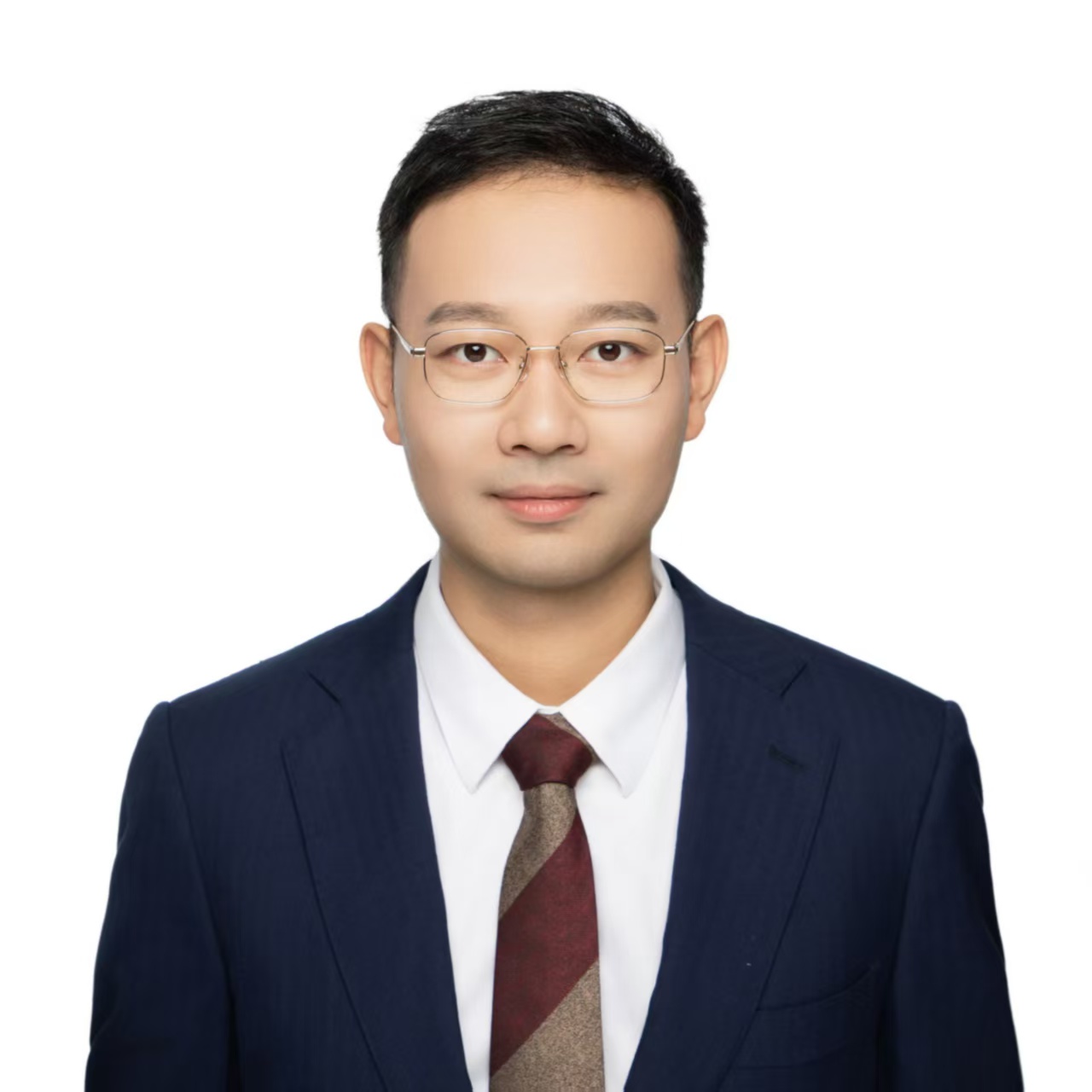}}]{Guo Chen} received the Ph.D. degree from Tsinghua University in 2016. He was a Researcher at Microsoft Research Asia from 2016 to 2018 and the Deputy Director of the Office of Informatization at Hunan University from 2020 to 2023. He is currently a Professor at Hunan University. His long-term research focuses on high-performance networks, cloud computing networking, and AI systems. His work on RDMA protocol stacks and routing transmission protocols has been applied in top companies such as Huawei, Tencent,  Baidu and ByteDance.
\end{IEEEbiography}

\end{document}